\documentclass[10pt,twocolumn,letterpaper]{article}

\usepackage{iccv}              

\definecolor{iccvblue}{rgb}{0.21,0.49,0.74}
\usepackage[pagebackref,breaklinks,colorlinks,allcolors=iccvblue]{hyperref}
\usepackage{graphicx} 
\usepackage{subcaption} 
\usepackage[margin=1in]{geometry} 
\usepackage{multirow}

\def\paperID{8} 
\def\confName{ICCV}
\def\confYear{2025}
\newcommand{\method}{DeQA-Doc\xspace}

\title{DeQA-Doc: Adapting DeQA-Score to Document Image Quality Assessment}

\author{
\hspace{-11pt}
Junjie Gao$^{1,5}$, Runze Liu$^2$, Yingzhe Peng$^2$, Shujian Yang$^{1,3}$, Jin Zhang$^1$, Kai Yang$^{1\dag}$, Zhiyuan You$^{4\dag}$ \vspace{2pt} \\
\hspace{-10pt}
$^1$Ant Group, 
$^2$Southeast University, 
$^3$Shanghai Jiao Tong University, 
$^4$CUHK,
$^5$MBZUAI\vspace{2pt}\\
{\hspace{-16pt}
\tt\small junjiegao19@gmail.com, \{runze.liu, yingzhe.peng\}@seu.edu.cn, thomasyang0925@sjtu.edu.cn} \\
{\tt\small \{zj346862, chengyue.yk\}@antgroup.com, zhiyuanyou@foxmail.com} \quad 
{\small $^\dag$ Corresponding Author}
}

\begin{document}
\maketitle
\begin{abstract}

Document quality assessment is critical for a wide range of applications including document digitization, OCR, and archival. 
However, existing approaches often struggle to provide accurate and robust quality scores, limiting their applicability in practical scenarios. 
With the rapid progress in Multi-modal Large Language Models (MLLMs), recent MLLM-based methods have achieved remarkable performance in image quality assessment. 
In this work, we extend this success to the document domain by adapting DeQA-Score, a state-of-the-art MLLM-based image quality scorer, for document quality assessment. 
We propose \method, a framework that leverages the visual language capabilities of MLLMs and a soft label strategy to regress continuous document quality scores. 
To adapt DeQA-Score to \method, we adopt two complementary solutions to construct soft labels without the variance information. 
Also, we relax the resolution constrains to support the large resolution of document images. 
Finally, we introduce ensemble methods to further enhance the performance. 
Extensive experiments demonstrate that \method significantly outperforms existing baselines, offering accurate and generalizable document quality assessment across diverse degradation types. 
Codes and model weights are available in \url{https://github.com/Junjie-Gao19/DeQA-Doc}. 

\end{abstract}

\section{Introduction}\label{sec:intro}

Assessing the quality of document images is a fundamental task in document understanding~\cite{diqa_survey, diqa_deep, cg_diqa}, with wide-ranging applications in digitization, OCR, information retrieval, and archival systems~\cite{min2017unified, schulz2022identity, shemiakina2021method}. 
Traditional Document Image Quality Assessment (DIQA) methods rely heavily on hand-crafted features or shallow models~\cite{cg_diqa, Kumar2012Sharpness, alaei2019new, alaei2015document, xu2016no, diqa_survey}, which often fail to capture high-level semantic and structural information. While more recent approaches incorporate deep learning-based metrics~\cite{diqa_deep,lu2019deep, alaei2023document, zhang2024efficient} but still struggle to generalize across diverse document layouts and degradation types. In contrast, Multi-modal Large Language Models (MLLMs) have shown promising capabilities in related fields such as Image Quality Assessment (IQA)~\cite{liqe, qinstruct, coinstruct, depictqa, depictqav2, qalign, compare2score, deqa_score}, owing to their ability to jointly reason over visual and textual modalities.

Building on this progress, we explore the application of MLLM-based quality prediction to the document domain. Specifically, we adapt DeQA-Score~\cite{deqa_score}, a state-of-the-art MLLM framework, originally developed for image quality scoring to perform document quality assessment. We introduce \method, a simple yet effective extension that leverages MLLMs' visual-linguistic alignment and incorporates a distribution-based soft label strategy to regress continuous quality scores.

A key challenge in adapting DeQA-Score to documents lies in the lack of annotation variance information in DIQA datasets, which is required for constructing soft labels. 
To address this, we propose two complementary solutions: (1) assigning a fixed pseudo variance based on empirical statistics from other datasets, and (2) using linear interpolation between adjacent quality levels as a lightweight surrogate for cases where only mean opinion scores are available.

Another challenge lays in that document images often have significantly larger resolutions, but DeQA-Score can only handle images with a fixed resolution of $448 \times 448$. 
DeQA-Score uses mPLUG-Owl2 as the base model, and its vision encoder is implemented with a CLIP backbone~\cite{clip}, which by default only accepts input images of a fixed resolution, typically $448 \times 448$. 
However, document images often have significantly larger resolutions, and aggressive down-sampling may lead to severe information loss, especially for fine-grained layout and text details. 
To enable flexible input resolutions, we propose two alternative methods. 
First, we can remove the absolute position embeddings from the CLIP encoder. 
Second, alternatively, we can change the base model from mPLUG-Owl2 to Qwen2.5-VL, which can take dynamic and original image resolutions. 
These modifications decouple the model from the original fixed spatial grid and allow it to process high-resolution document images directly, preserving more structural and semantic information during encoding.

Our framework is instantiated on top of two strong MLLM backbones, mPLUG-Owl2~\cite{mplugowl2} and Qwen2.5-VL~\cite{qwen2.5vl}, and incorporates architectural modifications such as resolution-flexible vision encoding to better accommodate high-resolution document images. 
We adopt a hybrid training objective that combines cross entropy loss for next token prediction and KL divergence loss for soft label supervision. 
To enhance performance, we further apply two ensemble strategies: model ensemble, which averages predictions from mPLUG-Owl2 and Qwen2.5-VL, and prompt ensemble, which aggregates outputs from multiple semantically diverse prompts.
These strategies effectively leverage complementary model capacities and mitigate prompt sensitivity, yielding more robust and accurate predictions, as validated in our experiments.

Extensive experiments demonstrate that \method achieves competitive performance on the DIQA-5000 dataset, significantly outperforming existing baselines. The proposed method offers a robust generalizable solution for document quality assessment, and highlights the potential of MLLMs to unify image and document quality evaluation under a shared framework.

\section{Related Works}

\textbf{Document Image Quality Assessment} (DIQA) aims to automatically evaluate the quality of a document image. 
In this context, quality is often synonymous with readability, which can be assessed from either a machine perspective, typically measured by OCR accuracy, or a human perspective, based on perceptual visual quality. 
Traditional DIQA methods rely heavily on hand-crafted features or shallow models~\cite{alaei2015document, cg_diqa, Kumar2012Sharpness, alaei2019new, xu2016no, diqa_survey}. 
Deep-learning-based methods train models on large-scale datasets~\cite{diqa_deep, lu2019deep, alaei2023document}, significantly improving the performance. 
However, existing approaches often struggle to provide accurate and robust quality scores, limiting their applicability in practical scenarios.

\vspace{2pt}\noindent\textbf{Image Quality Assessment} (IQA) methods are categorized into Full-Reference (FR) and No-Reference (NR) based on the availability of a pristine reference image. 
FR-IQA computes a similarity score to a reference. Classical methods rely on hand-crafted metrics like structural similarity~\cite{ssim} and phase congruency with gradient magnitude~\cite{fsim}. 
Driven by large-scale datasets~\cite{lpips, koniq, pipal}, deep learning-based approaches~\cite{dists, A-DISTS, ghildyal2022stlpips, afine}, pioneered by LPIPS~\cite{lpips} and PieAPP~\cite{pieapp}, have achieved high accuracy in quality regression. 
NR-IQA directly predicts a quality score. Early works used hand-crafted natural image statistics~\cite{niqe, ma2017learning, saad2012blind}. 
Recent advances further improve accuracy through strategies like multi-dataset co-training~\cite{unique}, multi-scale features~\cite{musiq}, CLIP pre-training~\cite{clipiqa}, multi-dimension attention~\cite{maniqa}, and multitask learning~\cite{liqe}.

\vspace{2pt}\noindent\textbf{MLLM-based IQA} methods leverage MLLMs' powerful foundational knowledge to achieve better performance and detailed assessment results~\cite{iqasurvey_tianhe}. Researches have focus on low-level perception ability~\cite{qinstruct}, multi-image quality comparison~\cite{coinstruct}, and fine-grained visual quality grounding and analysis~\cite{qground}. 
Some works have proposed quality scoring strategies. 
Q-Bench~\cite{qbench} and Q-Instruct~\cite{qinstruct} propose a binary softmax strategy, enabling MLLMs to generate quality scores by predicting two discrete quality levels. 
Inspired by humans' annotation process, Q-Align~\cite{qalign} discretizes scores to five discrete levels using one-hot label to train MLLMs. 
DeQA-Score~\cite{deqa_score} introduces score distribution regression, further improving the performance.

\section{Method}\label{sec:method}

\subsection{Soft Label Construction}\label{sec:soft}

DeQA-Score~\cite{deqa_score} requires both the mean and variance of quality scores to construct the distribution-based soft label for training. 
However, the DIQA-5000 dataset only provides the mean scores without corresponding variances, making direct construction of soft labels infeasible. 
To address this limitation, we propose two alternative strategies. 
First, we introduce a pseudo-variance scheme by assigning a fixed variance value, allowing us to approximate the underlying score distribution and construct a soft label accordingly.
Second, we adopt a linear interpolation approach, which assumes the score lies between two adjacent discrete levels and assigns probability mass only to those two levels.

\begin{figure}[t]
    \centering
    \includegraphics[width=1.0\linewidth]{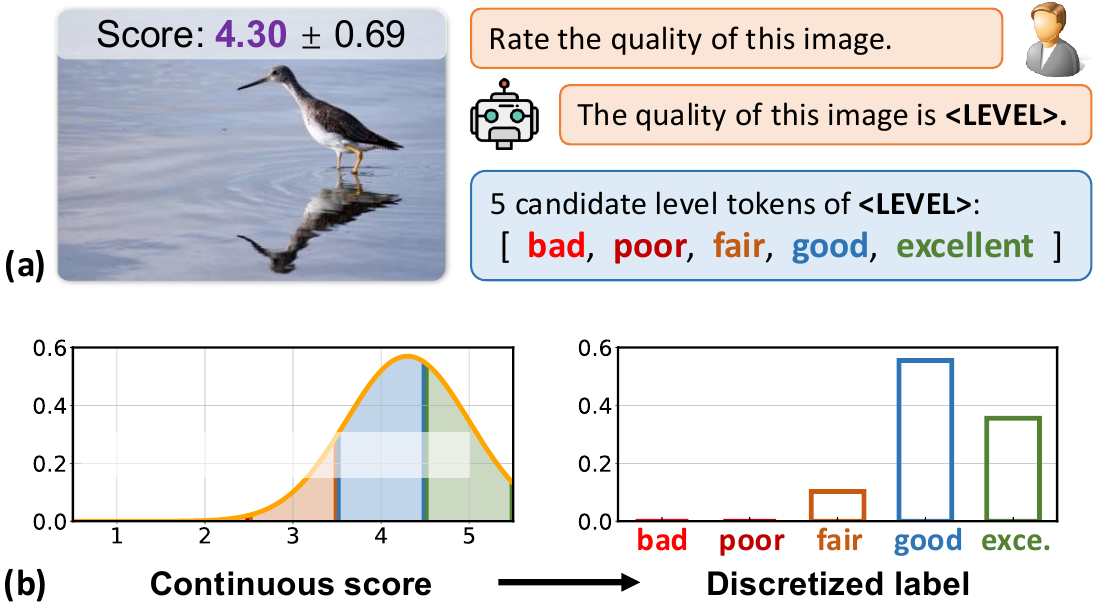}
    \caption{        
    (a) \textbf{MLLMs need discrete level tokens for training}. 
    Thus, continuous scores must be converted into discrete level tokens as training labels.
    (b) \textbf{Soft label in DeQA-Score~\cite{deqa_score}}. 
    DeQA-Score discretizes the estimated Gaussian distribution of the score. 
    The soft label better preserves the relationships between levels, \eg, ``fair'' is closer to ``good'' than ``excellent''. 
    This figure is adapted from DeQA-Score~\cite{deqa_score}. 
    }
    \vspace{-5pt}
    \label{fig:soft}
\end{figure}

\subsubsection{Soft Label in DeQA-Score}

\textbf{Discretization using score distribution}. 
MLLMs only accept discrete tokens as inputs and outputs.
Therefore, as shown in \cref{fig:soft}a, continuous scores\footnote{Unless otherwise specified, the mean quality scores will be normalized to $[1, 5]$, and the variances will be normalized accordingly.} must be discretized into discrete tokens to train the MLLMs. 
There are five rating levels, \{``bad'', ``poor'', ``fair'', ``good'', ``excellent''\}, which is the rating standard defined by \cite{five_level}. 
As demonstrated in~\cite{unique}, the quality score of one image, $x$, is typically modeled as a Gaussian distribution, $x \sim \mathcal{N}(\mu, \sigma^2)$, where $\mu$ and $\sigma^2$ are the Mean Opinion Score (MOS) and variance of multiple human annotations, respectively. 
The distribution is represented by its probability density function, $f(x)$. 
As shown in \cref{fig:soft}b, DeQA-Score selects five central points, $c_i \in \{1,2,3,4,5\}, i \in \{0,1,2,3,4\}$, as the centers of the five discrete levels. 
The width of each level region, $d$, is set as 1. 
The probability of the score falling into the $i^{\rm th}$ level is denoted as $p_i^{raw}$ and is calculated as: 
\begin{equation}
    p_i^{raw} = \int_{c_i - d/2}^{c_i + d/2}f(x){\rm dx}, \quad i \in \{0,1,2,3,4\}.
    \label{eq:prob_level}
\end{equation}
The five levels are assigned with the textual tokens including \{``bad'', ``poor'', ``fair'', ``good'', ``excellent''\}. 
Correspondingly, the score of $i^{\rm th}$ level is $c_i$. 
This defines a rough discrete distribution as $P^{raw}(c_i) = p_i^{raw}$.

\vspace{2pt}\noindent\textbf{Post-adjustment}. 
The discrete distribution still has truncation errors. 
As shown in \cref{fig:soft}b, there are two truncation regions (\ie, $x < c_0 - d / 2 = 0.5$ and $x > c_4 + d / 2=5.5$), which are not considered in \cref{eq:prob_level}. 
This can cause two problems. 
First, the sum of $\{p_i^{raw}\}$ does not equal 1, meaning $p_i^{raw}$ is not strictly a discrete distribution. 
Second, the expectation of this discrete distribution does not equal the expectation of score distribution, $\mu$, leading to discretization errors.

\vspace{2pt}\noindent To correct these truncation errors, DeQA-Score proposes a post-adjustment method. 
Specifically, DeQA-Score applies a linear transformation to adjust $p_i^{raw}$ to more accurate $p_i$ as: 
\begin{align}
    p_i &= \alpha p_i^{raw} + \beta \nonumber \\
    {\rm s.t.} \quad \sum\nolimits_i p_i &= 1, \quad \mu^{rec} = \sum\nolimits_i p_i c_i = \mu, \label{eq:adjust}
\end{align}
where $\mu_{rec}$ means recovered mean from discrete distribution. 
The parameters of the linear transformation, $\alpha$ and $\beta$, can be determined by solving these two constraint conditions.
This effectively defines the soft label, which follows the discrete distribution as $P(c_i) = p_i$.

\subsubsection{Constructing Soft Label without Variance}

The DIQA-5000 dataset provides only Mean Opinion Scores (MOS) without variance, making Gaussian-based soft label construction infeasible.
To address this, we adopt two alternatives: (1) assign a fixed pseudo-variance to approximate the score distribution, and (2) apply linear interpolation between two adjacent levels to form a two-level soft label.

\vspace{2pt}\noindent\textbf{Direct use of pseudo variance due to missing variance}. 
The DIQA-5000 dataset only provides Mean Opinion Scores (MOS) without variance annotations, which prevents the direct construction of soft labels based on Gaussian score distributions. 
To address this, we assign a pseudo variance to each sample, following the empirical statistics observed from other IQA datasets. 
As shown in DeQA-Score~\cite{deqa_score}, the average standard deviation across datasets such as KonIQ~\cite{koniq}, SPAQ~\cite{spaq}, and KADID~\cite{kadid} typically accounts for around 20\% of the full score range. 
Based on this observation, we set the pseudo standard deviation in DIQA-5000 as $0.2 \times (\text{max} - \text{min})$, where \text{max} and \text{min} are the upper and lower bounds of the original score range. 
All scores are then normalized to [1, 5] during training, and the pseudo variance is scaled accordingly. 
This approximation enables soft label construction in the absence of ground-truth variance and allows training seamlessly on the DIQA-5000 dataset.

\vspace{2pt}\noindent\textbf{Direct use of linear interpolation due to missing variance}. 
As discussed in DeQA-Score~\cite{deqa_score}, when the variance is extremely small, the Gaussian distribution of the quality score can be approximated by a unit impulse function, and the soft label naturally degrades into a linear interpolation between the two nearest levels. 
This provides a theoretical justification for directly using linear interpolation when the variance is unknown. 
In the DIQA-5000 dataset, only the Mean Opinion Scores (MOS) are available, without accompanying variance annotations. 
Therefore, we directly apply the linear interpolation strategy as a substitute for soft label construction. 
Specifically, given a mean score $\mu$, we identify the two adjacent level centers $c_j$ and $c_{j+1}$ such that $c_j < \mu \leq c_{j+1}$, and compute the soft label as: 
\begin{equation}
p_i = 
\left\{
\begin{aligned}
&c_{j+1} - \mu, & \text{if } i = j, \\
&\mu - c_j, & \text{if } i = j+1, \\
&0, & \text{otherwise}.
\end{aligned}
\right.
\end{equation}
This approximation is consistent with the behavior of our framework under very small variance, and serves as a simple yet effective surrogate when full distributional information is not available.

\subsection{Training with Soft Label}

\begin{figure*}[t]
    \centering
    \includegraphics[width=0.95\linewidth]{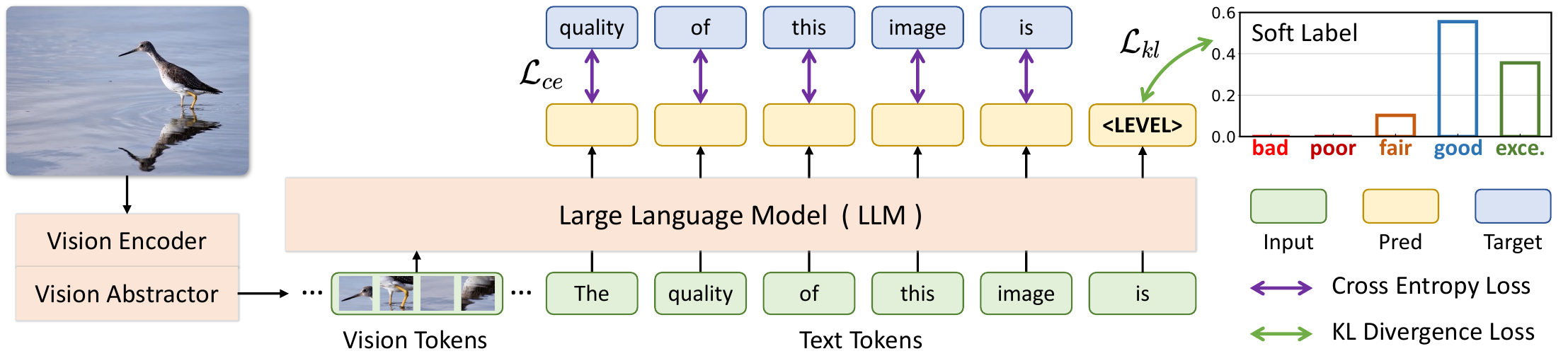}
    \caption{
    \textbf{Framework of our \method trained with soft label}. 
    For the ``\(<\)level\(>\)'' token, the KL divergence loss is calculated between predicted probabilities and our soft label. For other tokens, common cross-entropy loss for next token prediction is calculated. 
    This figure is borrowed from DeQA-Score~\cite{deqa_score}. 
    }
    \label{fig:model}
    \vspace{-5pt}
\end{figure*}

As illustrated in \cref{fig:model}, we adopt the same training strategy as DeQA-Score~\cite{deqa_score} to train our model within the MLLM framework.

\vspace{2pt}\noindent\textbf{Model architecture}.
Following the standard MLLM design~\cite{llava, mplugowl2}, our model consists of a vision encoder, a vision-to-text connector, and a Large Language Model (LLM). 
The vision encoder first extracts visual tokens from the input image. 
These visual tokens are then projected into the textual embedding space via a vision-to-text connector, and then fused with textual tokens. 
Finally, the fused visual and textual tokens are fed into the LLM for response generation.
In our experiments, we test two categories of MLLMs including mPLUG-Owl2~\cite{mplugowl2} and Qwen2.5-VL~\cite{qwen2.5vl}.

\vspace{2pt}\noindent\textbf{Relaxing the resolution constraints of CLIP}. 
DeQA-Score directly utilizes mPLUG-Owl2 as the base model, whose vision encoder is implemented with a CLIP backbone~\cite{clip}, which by default only accepts input images of a fixed resolution, typically $448 \times 448$. 
However, document images often have significantly larger resolutions, and aggressive down-sampling may lead to severe information loss, especially for fine-grained layout and text details. 
To enable flexible input resolutions, we propose two alternative methods. 
First, we can remove the absolute position embeddings from the CLIP encoder. 
Second, alternatively, we can change the base model from mPLUG-Owl2 to Qwen2.5-VL, which can take dynamic and original image resolutions. 
These modifications decouple the model from the original fixed spatial grid and allow it to process high-resolution document images directly, preserving more structural and semantic information during encoding.

\vspace{2pt}\noindent\textbf{Next token loss for response generation}.
The model is trained to generate templated responses like ``The quality of this image is \(<\)level\(>\)''.
For all tokens in the response except the level token, we apply the standard next-token prediction loss, i.e., the cross-entropy loss $\mathcal{L}_{ce}$, following prior work~\cite{llama, gpt3.5}.

\vspace{2pt}\noindent\textbf{KL divergence loss for soft label supervision}.
The level token is drawn from five candidates: {``bad'', ``poor'', ``fair'', ``good'', ``excellent''}.
Given a soft label $p_i$ assigning probability to each level, and the model-predicted probability $p_i^{pred}$ from softmax over all vocabulary tokens, we compute the KL divergence loss as:
\begin{equation}
\mathcal{L}_{kl} = \sum\nolimits_i p_i \log \left( \frac{p_i^{pred}}{p_i} \right).
\end{equation}
Using the full vocabulary for softmax ensures suppression of unrelated tokens during training.

\vspace{2pt}\noindent\textbf{Combined loss function}.
Both $\mathcal{L}_{ce}$ and $\mathcal{L}_{kl}$ are auto-regressive losses and can be jointly optimized.
Since the response template is fixed, $\mathcal{L}_{ce}$ converges quickly, allowing the model to focus on learning accurate level token distributions via $\mathcal{L}_{kl}$.

\subsection{Score Estimation and Ensemble Inference}

\textbf{Score estimation}. 
At inference time, we derive the predicted score distribution from the model-generated probabilities over the five predefined levels, $p_i^{pred}$. 
We then compute the predicted mean score as: 
{{
\footnotesize
\begin{equation}
\mu^{pred} = \sum\nolimits_i p_i^{pred} c_i.
\label{eq:pred}
\end{equation}
}}
There are two options to obtain $p_i^{pred}$. 
First, the same as Q-Align~\cite{qalign}, we can adopt the closed-set softmax over the five level tokens, which filters out irrelevant textual tokens. 
Second, as proven in DeQA-Score~\cite{deqa_score}, the trained MLLM consistently assigns probability mass to the correct level set, making full-vocabulary and closed-set softmax nearly equivalent in practice.

\vspace{2pt}\noindent\textbf{Model ensemble to enhance performance}. 
To further boost the robustness and accuracy of our predictions, we adopt a simple yet effective model ensemble strategy.
Specifically, we utilize both mPLUG-Owl2-7B and Qwen2.5-VL as base models and average their predicted score distributions at inference time.
For each input image, we obtain the predicted probability distribution over the five quality levels from each model, and compute the final prediction by averaging these distributions element-wise.
This ensemble approach leverages the complementary strengths of different architectures, \eg, mPLUG-Owl2’s strong vision-language alignment and Qwen2.5-VL’s language modeling capabilities, and helps mitigate individual model biases. 
As shown in \cref{tab:ensemble}, the ensemble results consistently outperform each single model across all metrics, demonstrating improved stability and generalization without requiring additional fine-tuning.

\vspace{2pt}\noindent\textbf{Prompt ensemble to enhance performance}. 
In addition to model-level ensemble, we explore a prompt ensemble strategy to further improve prediction accuracy and robustness.
Given that MLLMs are sensitive to the phrasing and structure of prompts, we design 10 semantically equivalent question templates (\eg, ``How is the quality of this document?'', ``Please rate the visual quality of this image.''). 
At inference time, we feed each input image into the model with all prompt variants and obtain a set of predicted score distributions. 
We then average these distributions element-wise to produce the final prediction. 
This strategy helps reduce the variance caused by prompt phrasing and enables the model to make more consistent and reliable predictions. 
While prompt ensemble is often employed to mitigate MLLMs’ prompt sensitivity and enhance robustness, our findings in \cref{tab:ensemble} indicate that it did not yield noticeable performance gains over single-prompt inference for this quality assessment task. 
This suggests that merely aggregating responses from diverse prompts may not always effectively translate the underlying prompt sensitivity into improved performance, perhaps due to the nature of the task or the characteristics of the prompt set.

\section{Experiments}\label{sec:exp}

\subsection{Implementation Details}

We adopt mPLUG-Owl2-7B~\cite{mplugowl2} and Qwen2.5-VL-7B~\cite{qwen2.5vl} as our base models for score regression on the DIQA-5000 dataset. 
The models are trained under two fine-tuning regimes: full parameter tuning and parameter-efficient LoRA fine-tuning~\cite{lora}. 
There are also different choices for input resolution including $448 \times 448$, $1024 \times 1024$, and $1536 \times 1536$. 
The pretrained weights of base models are used for model initialization. 
Additionally, we also examine the impact of pretraining on external image quality datasets such as KonIQ~\cite{koniq}. 
All models are trained on 8 NVIDIA RTX A100 GPUs with AdamW~\cite{adamw} as the optimizer. 
The initial learning rate is set to 2e-5 and decays gradually using the cosine decay strategy. 
Our training protocol involved 3 epochs. 
For the $448 \times 448$ input resolution, a batch size of 64 is employed. 
Due to GPU memory constraints, the batch size is reduced to 8 for $1024 \times 1024$ resolution and further reduced to 2 for $1536 \times 1536$ resolution.

\subsection{Metrics}\label{subsec:metric}

We evaluate model performance by comparing predicted scores against the ground truth Mean Opinion Scores (MOS). 
Two widely adopted correlation metrics are used: the Pearson Linear Correlation Coefficient (PLCC) and the Spearman Rank-order Correlation Coefficient (SRCC). 
For each rating dimension, including overall quality, sharpness, and color fidelity, we compute the score as the average of PLCC and SRCC:
\begin{equation}
\begin{aligned}
    & s_{\text{overall}} = 0.5 \times (\text{SRCC}_{\text{overall}} + \text{PLCC}_{\text{overall}}), \\
    & s_{\text{sharpness}} = 0.5 \times (\text{SRCC}_{\text{sharpness}} + \text{PLCC}_{\text{sharpness}}), \\
    & s_{\text{color}} = 0.5 \times (\text{SRCC}_{\text{color}} + \text{PLCC}_{\text{color}}).
\end{aligned}
\end{equation}
The final score is then computed as a weighted sum of the three dimension-wise scores:
\begin{equation}
s_{\text{final}} = 0.5 \times s_{\text{overall}} + 0.25 \times s_{\text{sharpness}} + 0.25 \times s_{\text{color}}.
\end{equation}
These four score results, $s_{\text{overall}}, s_{\text{sharpness}}, s_{\text{color}}, s_{\text{final}}$, are obtained through submitting inferred quality scores to the official workshop website. Therefore, the SRCC and PLCC results are not available.

\begin{figure*}[htbp] 
    \centering 

    \begin{subfigure}[b]{0.195\textwidth} 
    \centering
    \includegraphics[width=\linewidth, height=3cm, keepaspectratio]{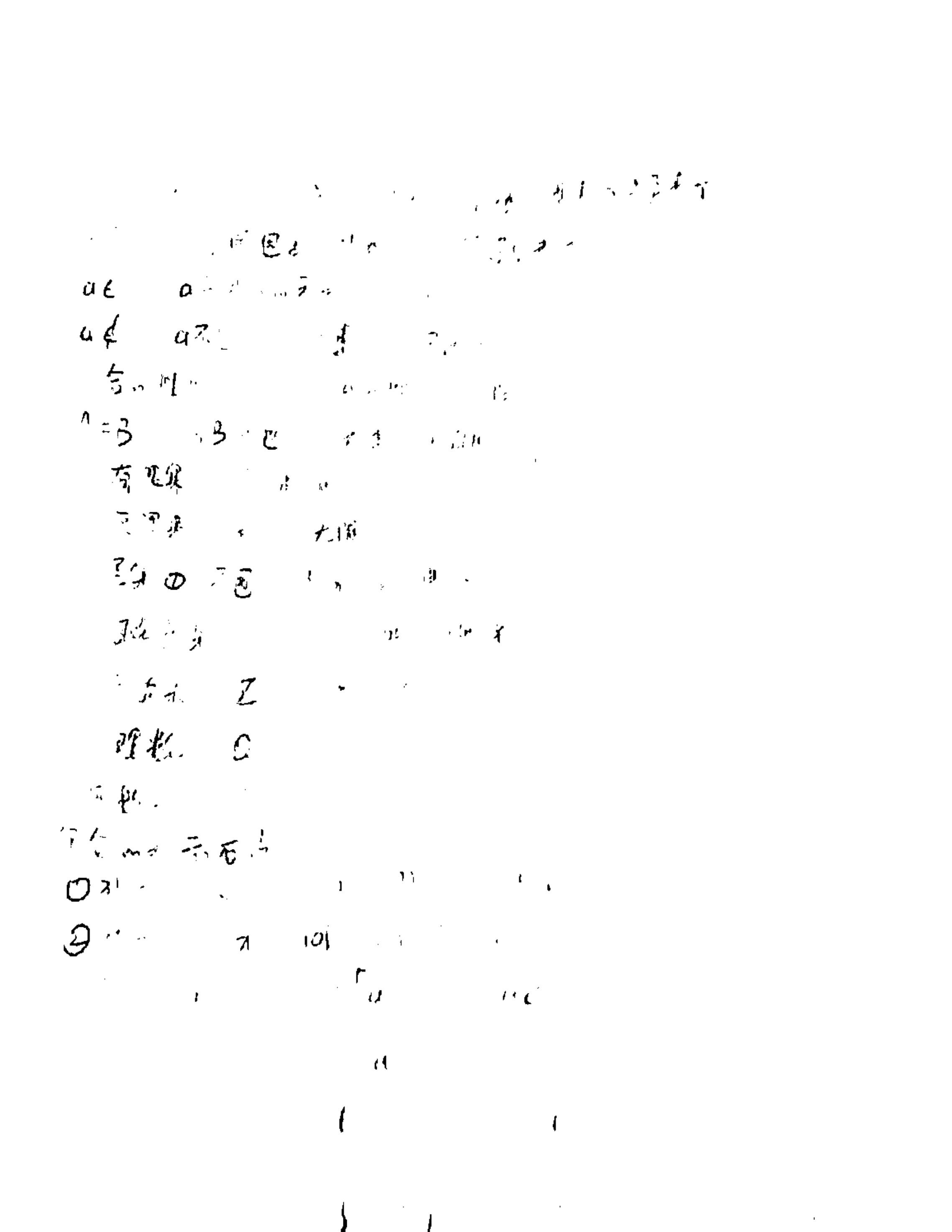} 
    \caption{$s_\text{overall}=1.145$} 
    \label{fig:image1} 
    \end{subfigure}
    \hfill 
    \begin{subfigure}[b]{0.195\textwidth}
    \centering
    \includegraphics[width=\linewidth, height=3cm, keepaspectratio]{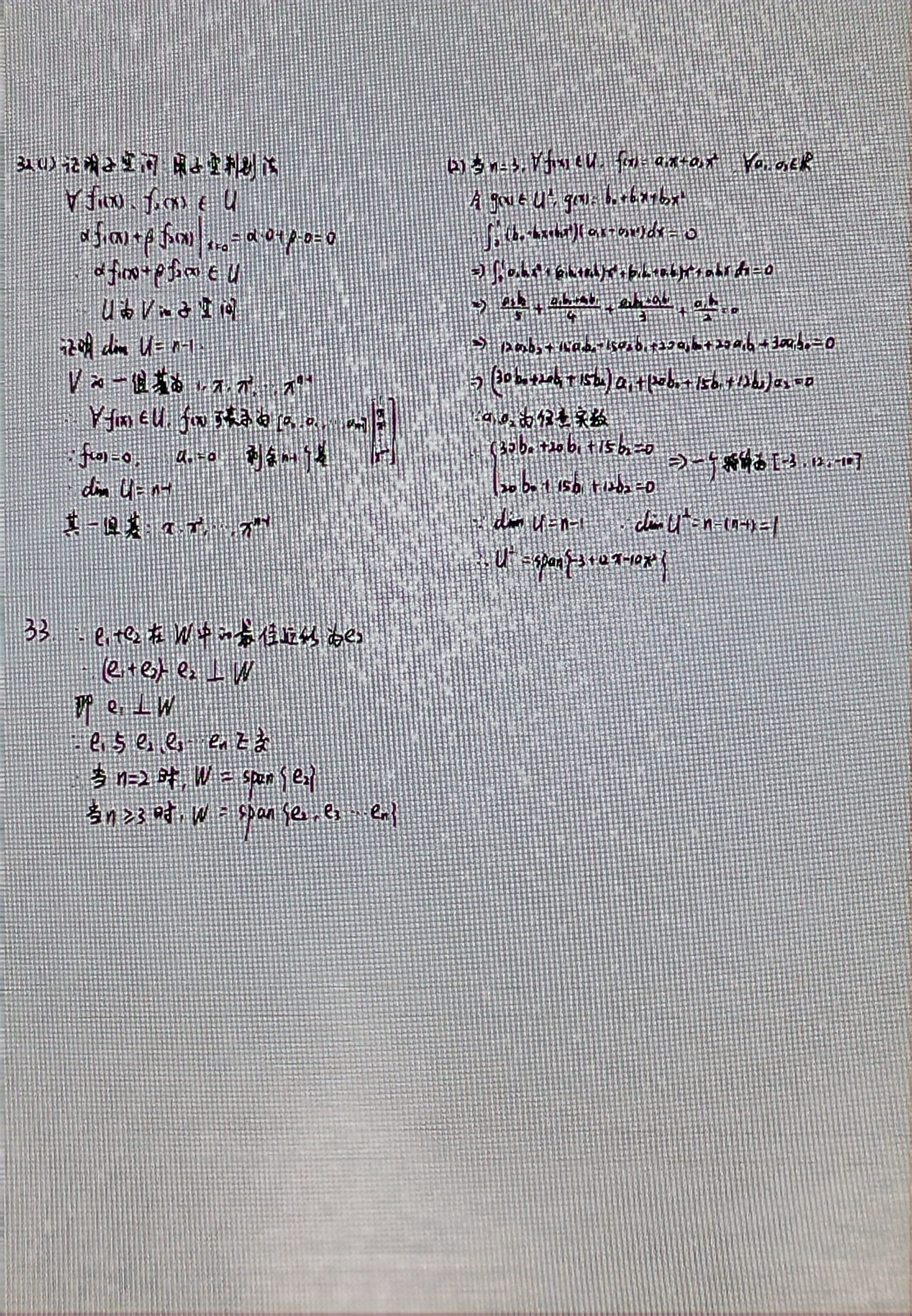}
    \caption{$s_\text{overall}=2.394$}
    \label{fig:image2}
    \end{subfigure}
    \hfill
    \begin{subfigure}[b]{0.195\textwidth}
    \centering
    \includegraphics[width=\linewidth, height=3cm, keepaspectratio]{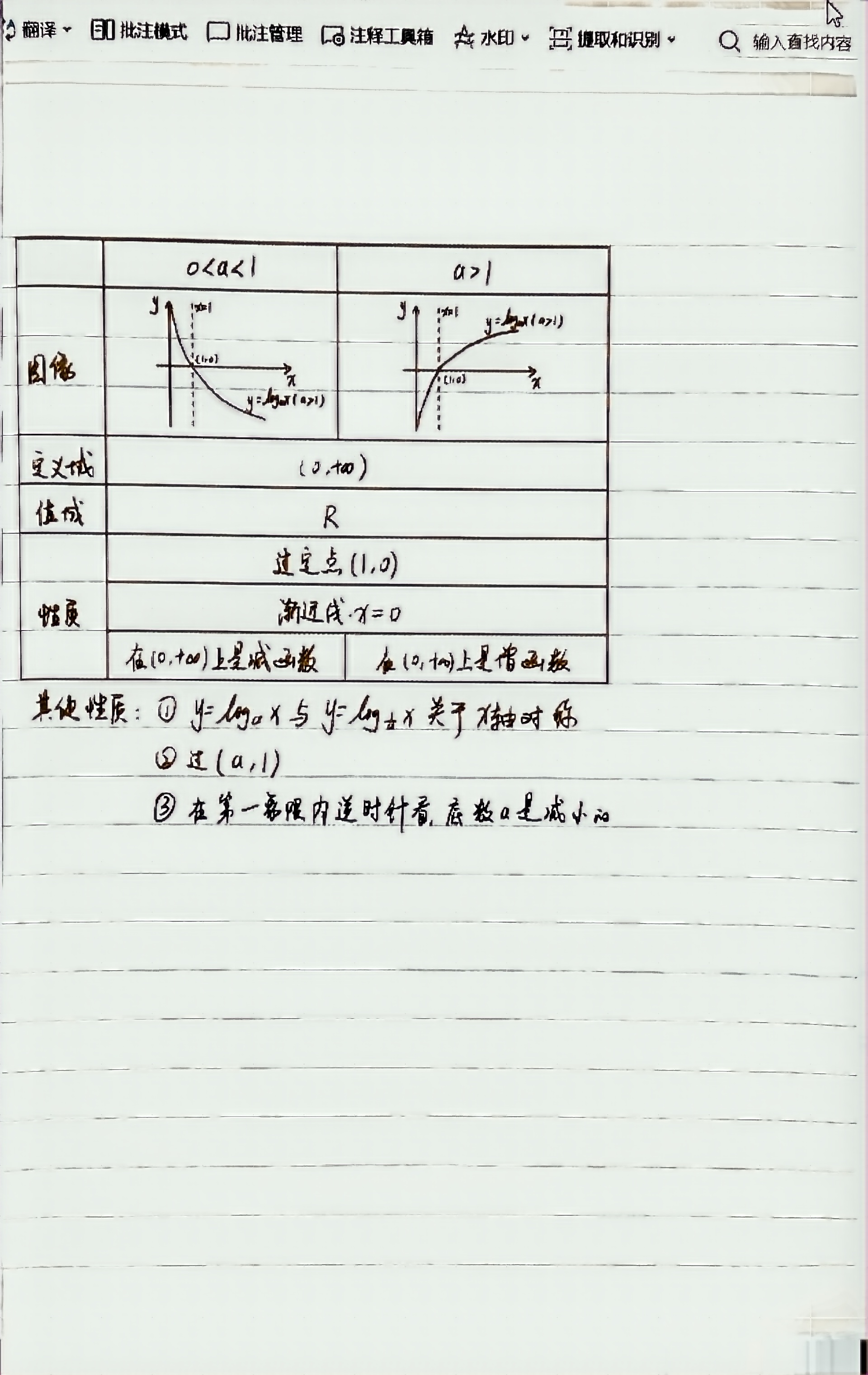}
    \caption{$s_\text{overall}=3.149$}
    \label{fig:image3}
    \end{subfigure}
    \hfill
    \begin{subfigure}[b]{0.195\textwidth}
    \centering
    \includegraphics[width=\linewidth, height=3cm, keepaspectratio]{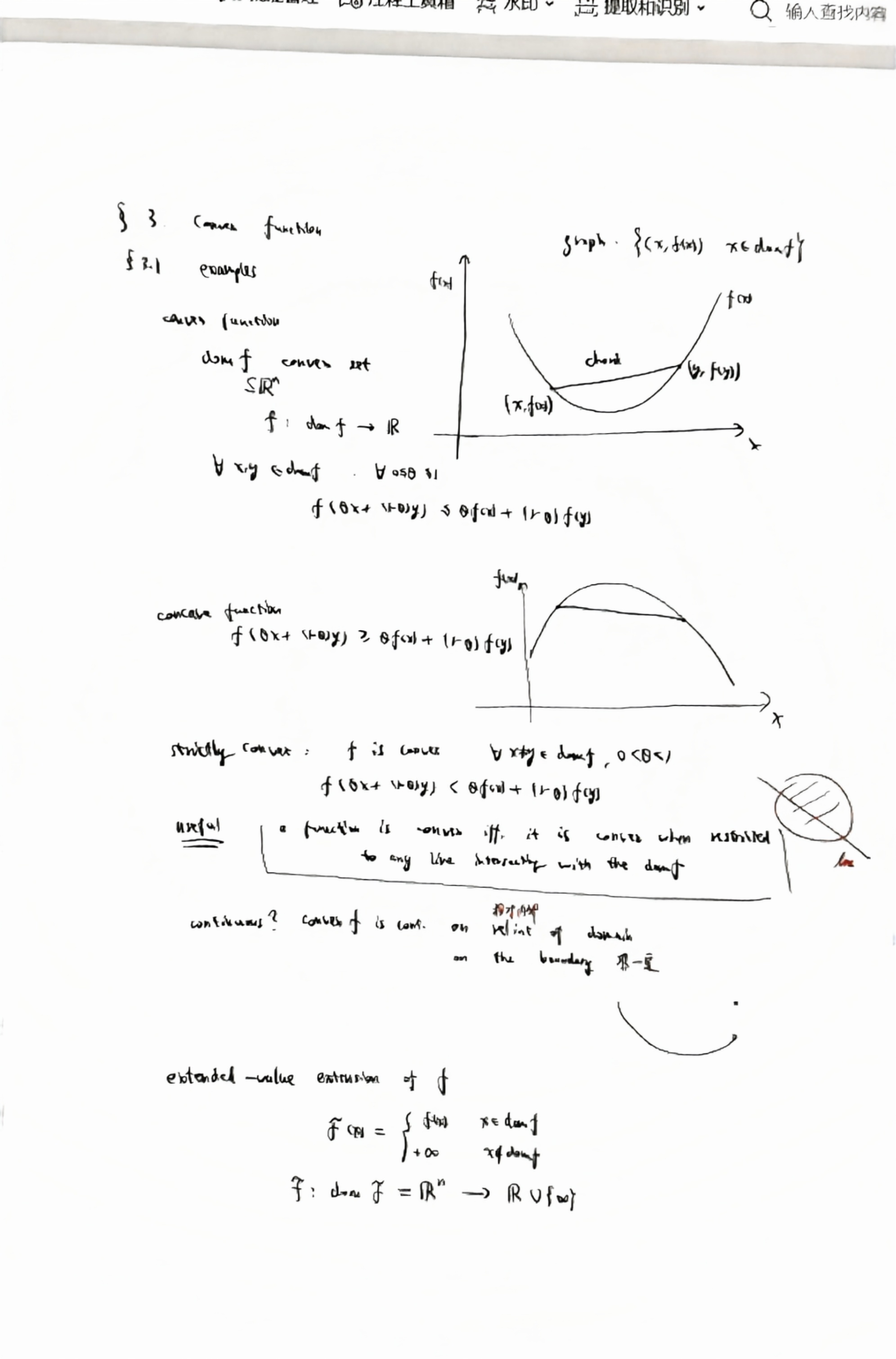}
    \caption{$s_\text{overall}=3.713$}
    \label{fig:image4}
    \end{subfigure}
    \hfill
    \begin{subfigure}[b]{0.195\textwidth}
    \centering
    \includegraphics[width=\linewidth, height=3cm, keepaspectratio]{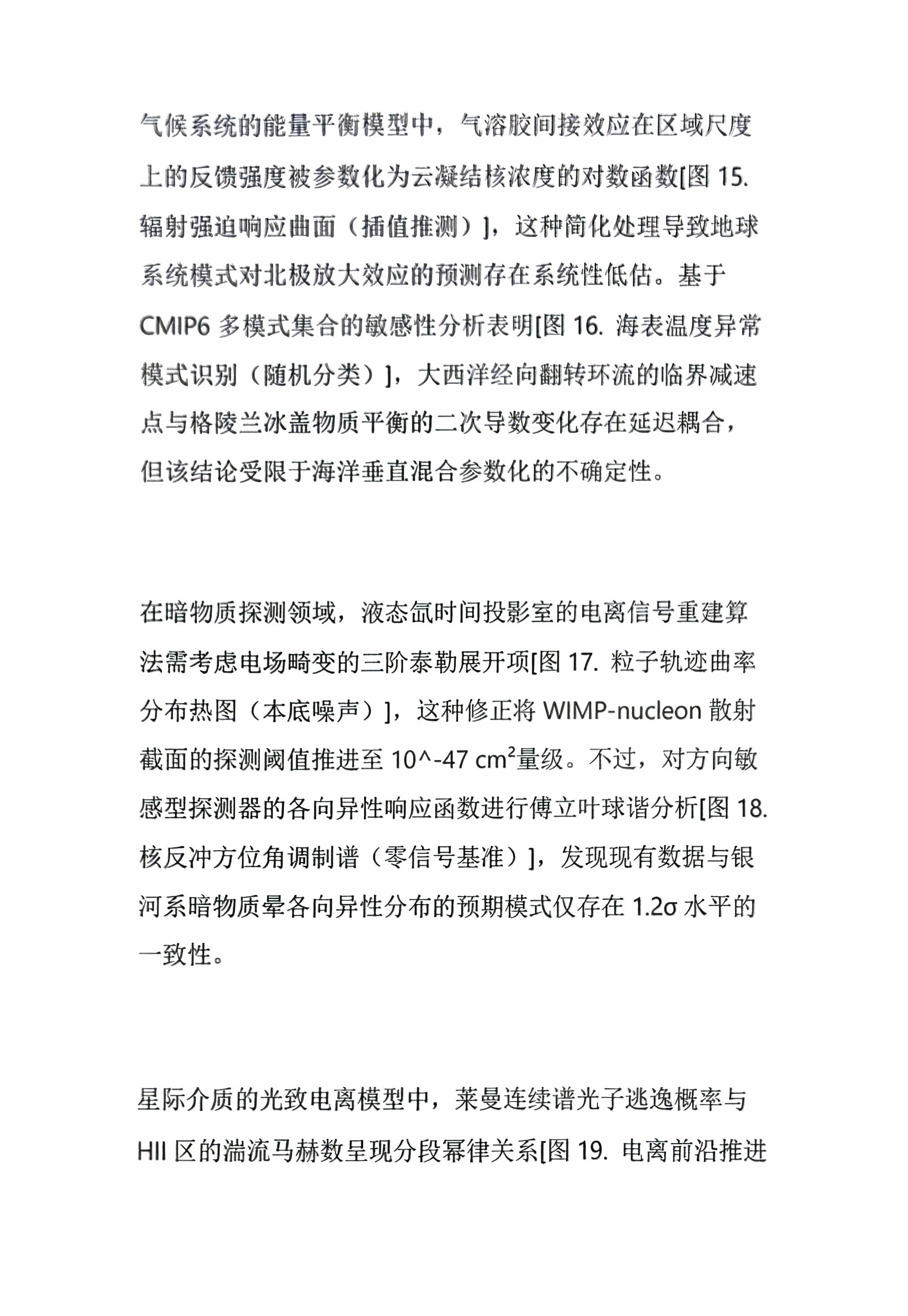}
    \caption{$s_\text{overall}=4.757$}
    \label{fig:image5}
    \end{subfigure}
    \par\bigskip 

    \begin{subfigure}[b]{0.195\textwidth}
    \centering
    \includegraphics[width=\linewidth, height=3cm, keepaspectratio]{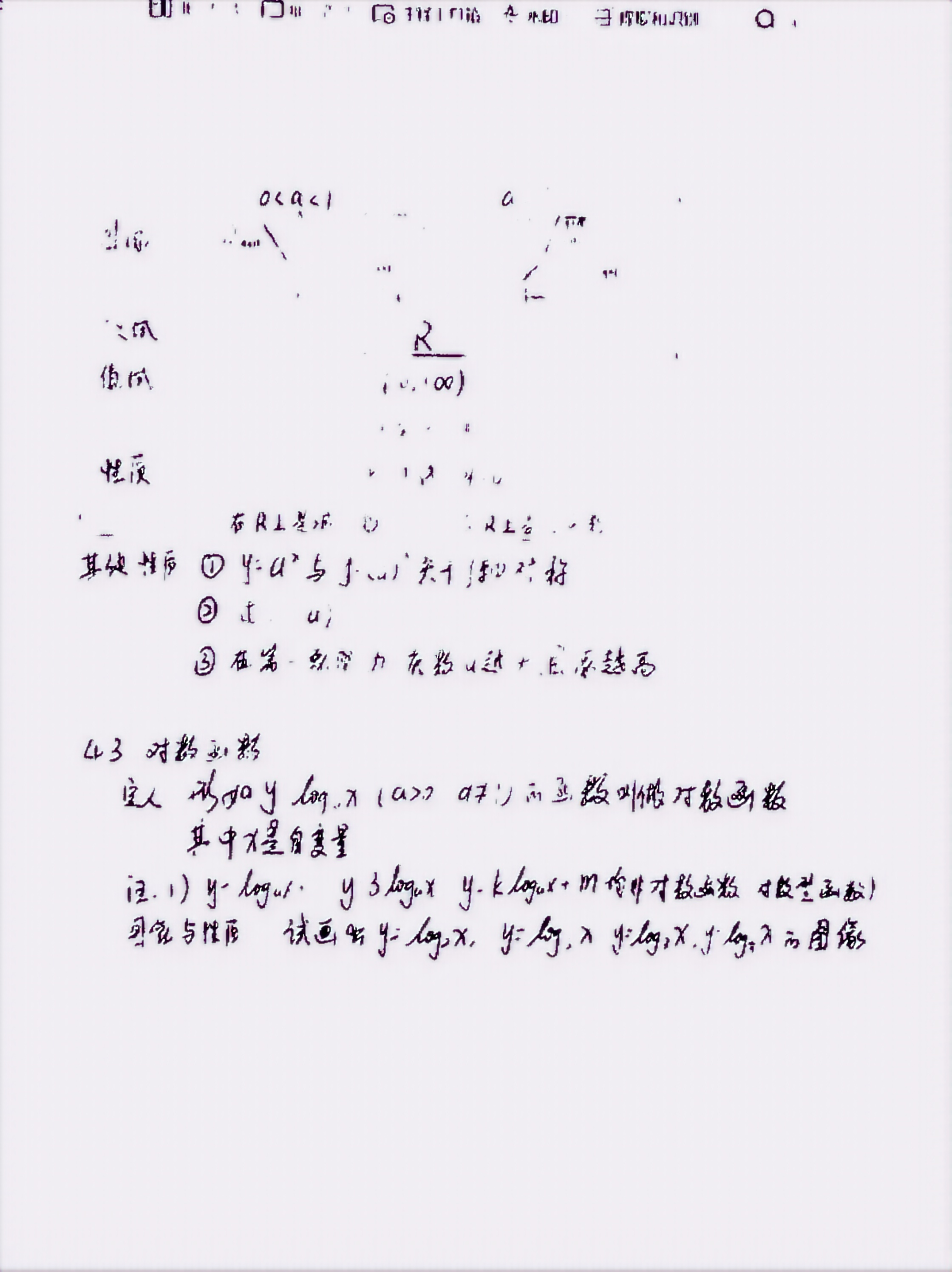}
    \caption{$s_\text{overall}=1.519$}
    \label{fig:image6}
    \end{subfigure}
    \hfill
    \begin{subfigure}[b]{0.195\textwidth}
    \centering
    \includegraphics[width=\linewidth, height=3cm, keepaspectratio]{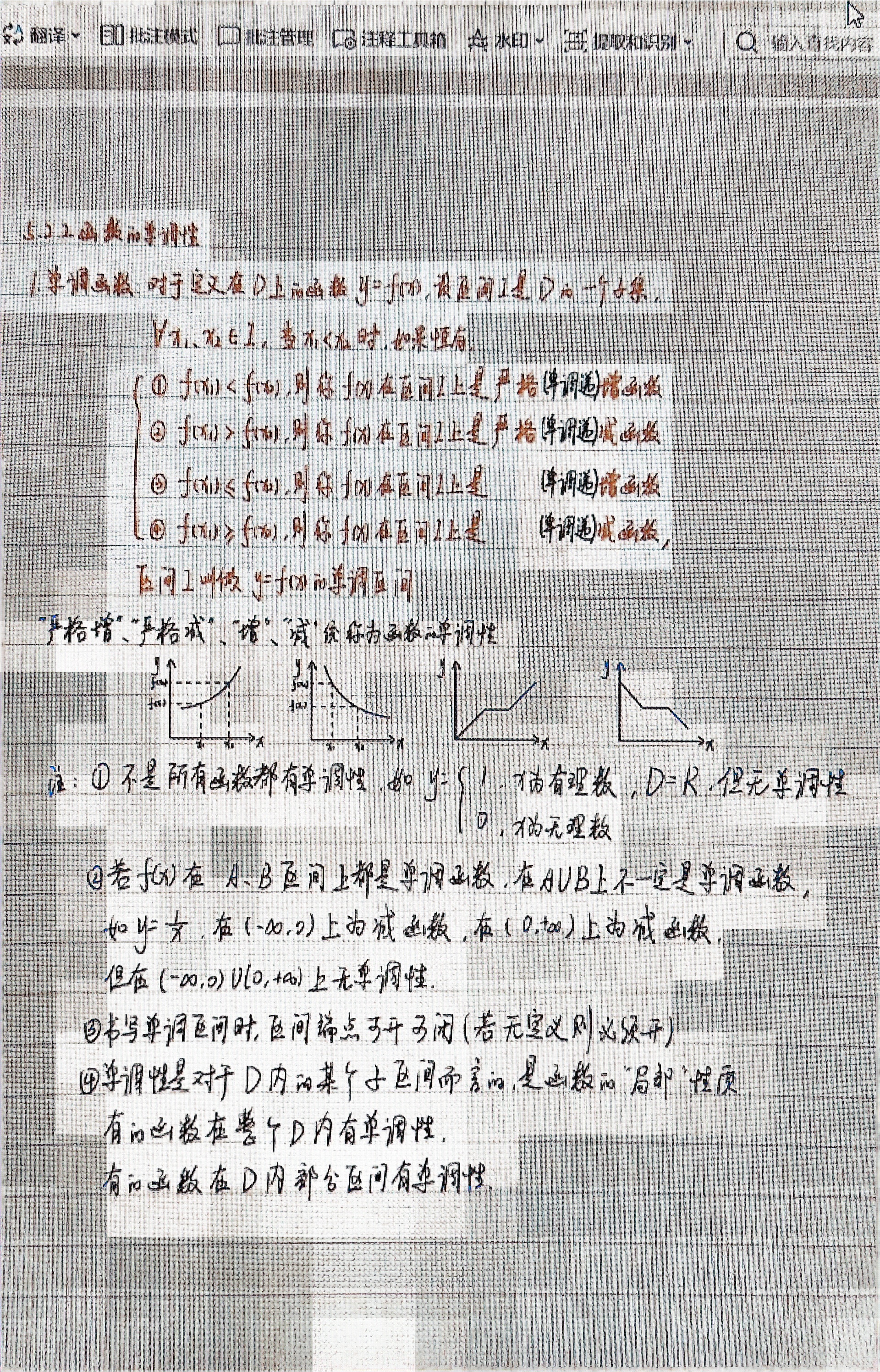}
    \caption{$s_\text{overall}=2.394$}
    \label{fig:image7}
    \end{subfigure}
    \hfill
    \begin{subfigure}[b]{0.195\textwidth}
    \centering
    \includegraphics[width=\linewidth, height=3cm, keepaspectratio]{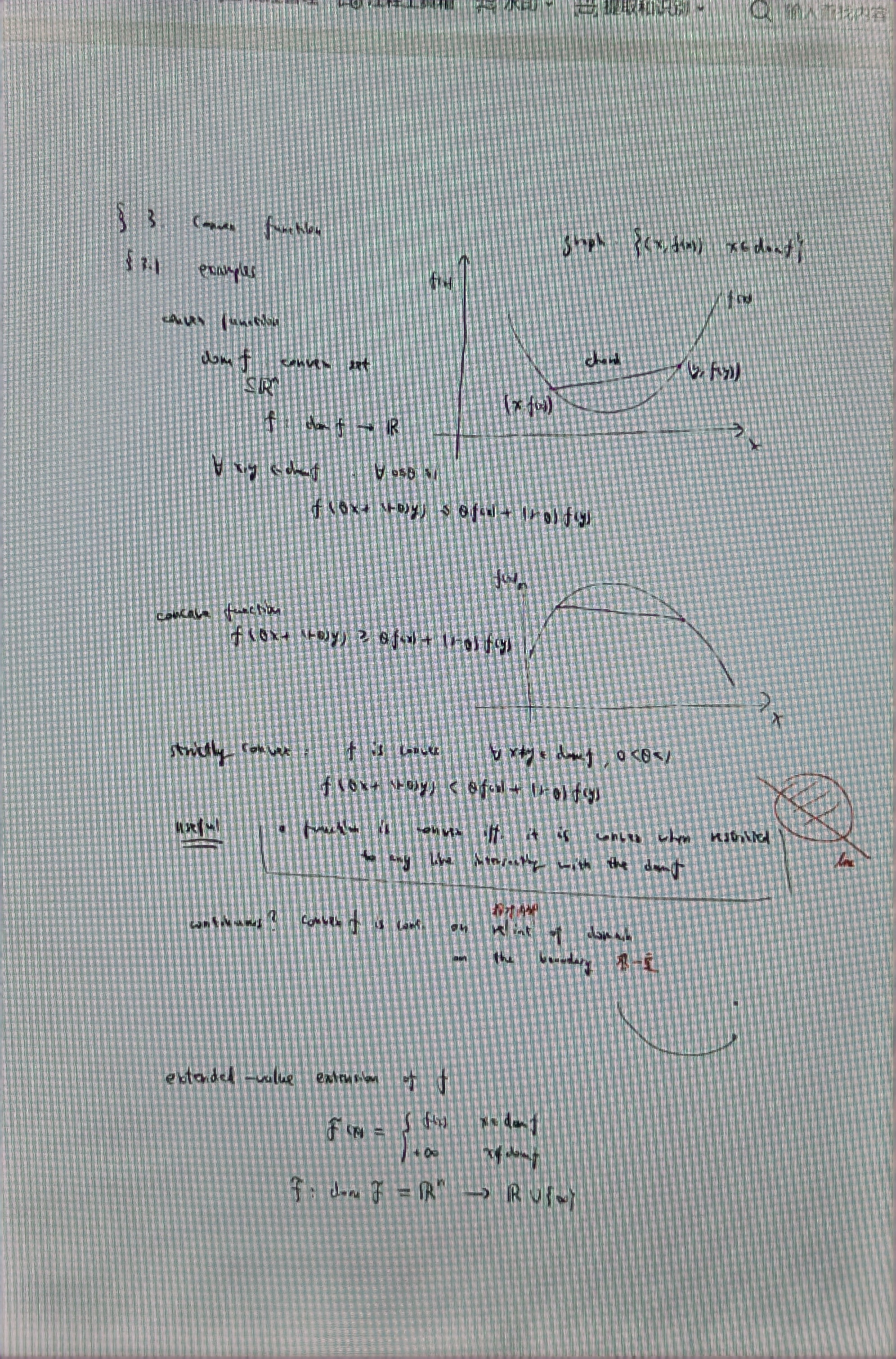}
    \caption{$s_\text{overall}=2.845$}
    \label{fig:image8}
    \end{subfigure}
    \hfill
    \begin{subfigure}[b]{0.195\textwidth}
    \centering
    \includegraphics[width=\linewidth, height=3cm, keepaspectratio]{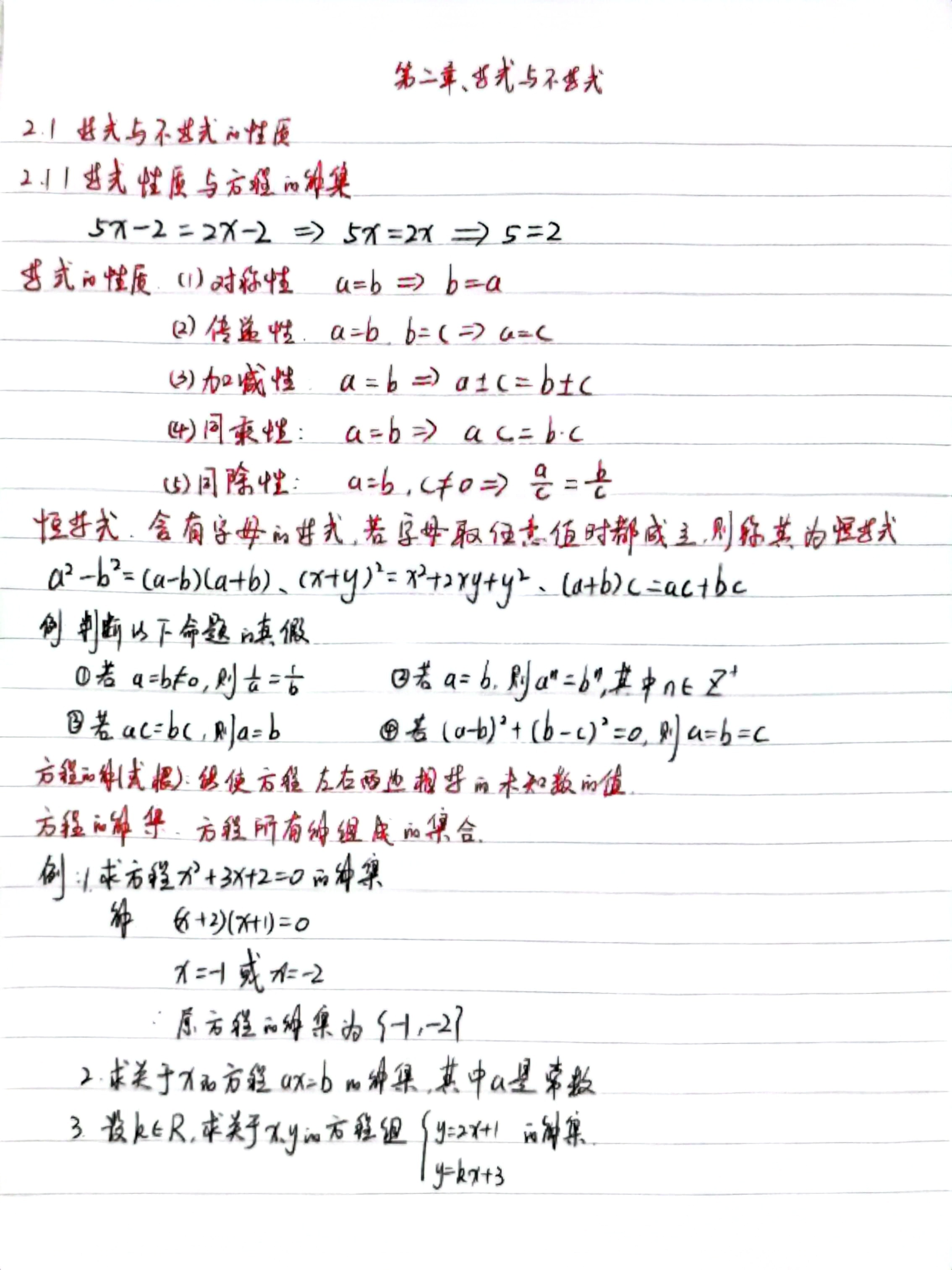}
    \caption{$s_\text{overall}=3.911$}
    \label{fig:image9}
    \end{subfigure}
    \hfill
    \begin{subfigure}[b]{0.195\textwidth}
    \centering
    \includegraphics[width=\linewidth, height=3cm, keepaspectratio]{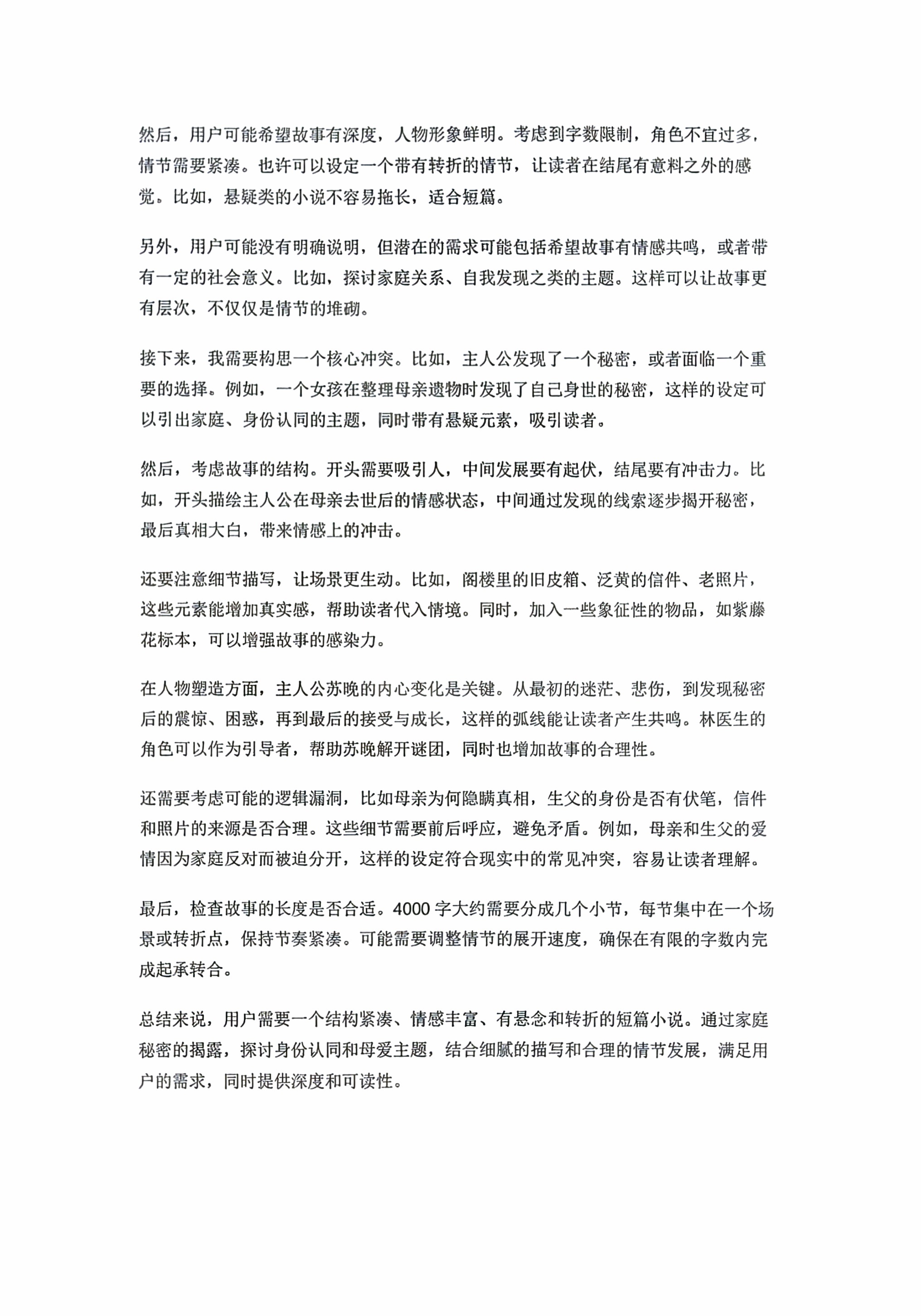}
    \caption{$s_\text{overall}=4.382$}
    \label{fig:image10}
    \end{subfigure}
    \par\bigskip 

    \begin{subfigure}[b]{0.195\textwidth}
    \centering
    \includegraphics[width=\linewidth, height=3cm, keepaspectratio]{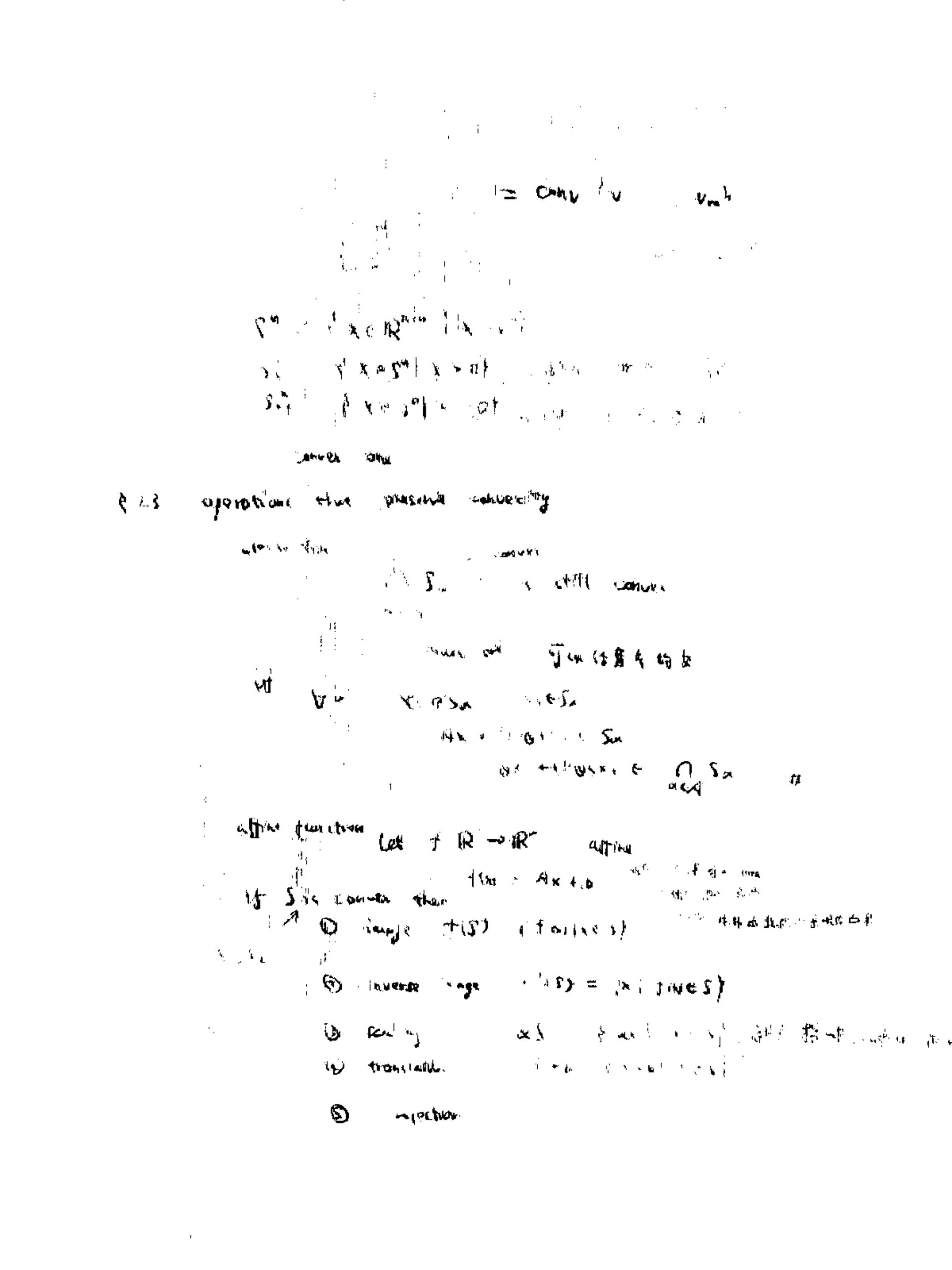}
    \caption{$s_\text{overall}=1.534$}
    \label{fig:image11}
    \end{subfigure}
    \hfill
    \begin{subfigure}[b]{0.195\textwidth}
    \centering
    \includegraphics[width=\linewidth, height=3cm, keepaspectratio]{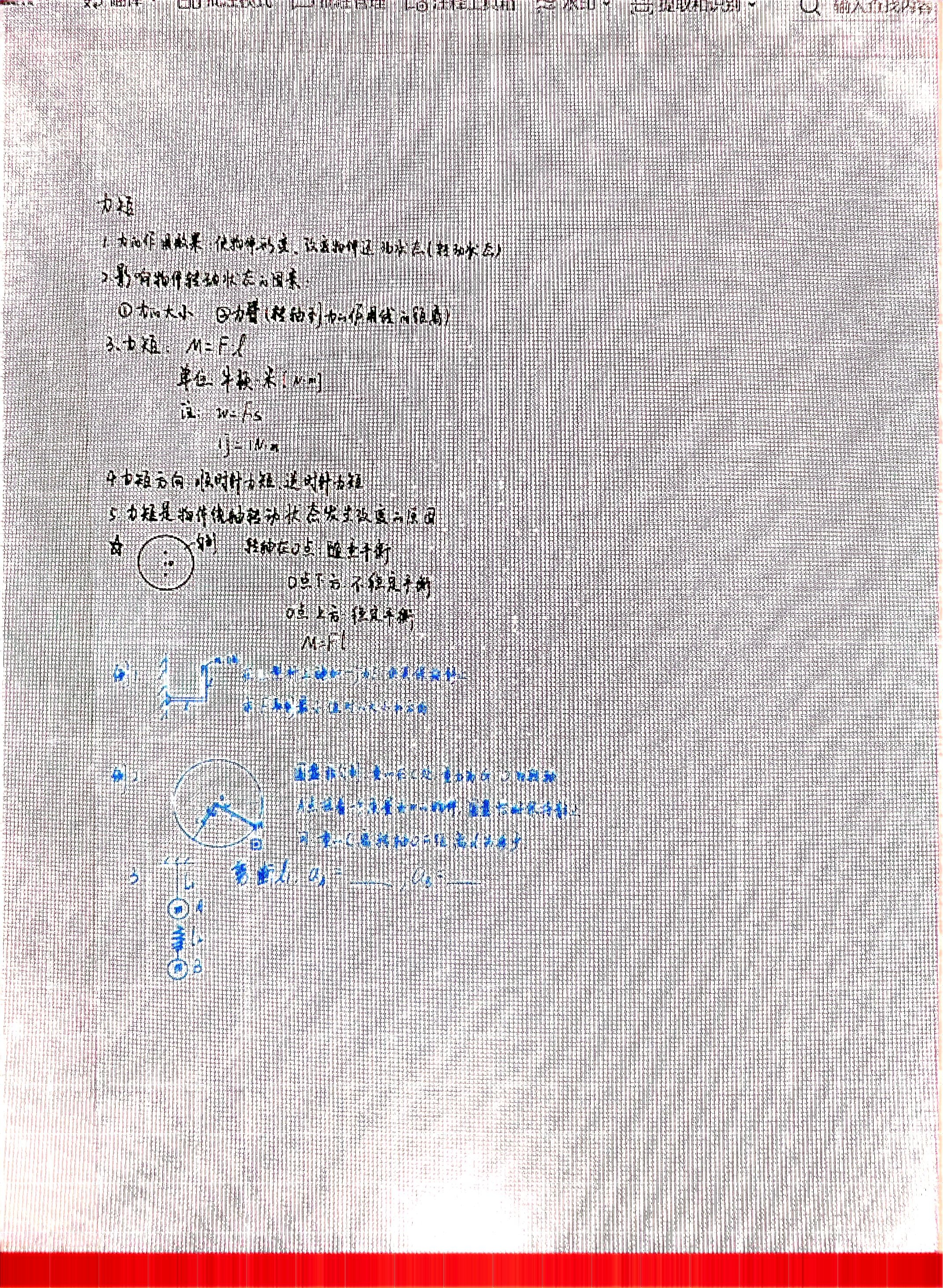}
    \caption{$s_\text{overall}=2.174$}
    \label{fig:image12}
    \end{subfigure}
    \hfill
    \begin{subfigure}[b]{0.195\textwidth}
    \centering
    \includegraphics[width=\linewidth, height=3cm, keepaspectratio]{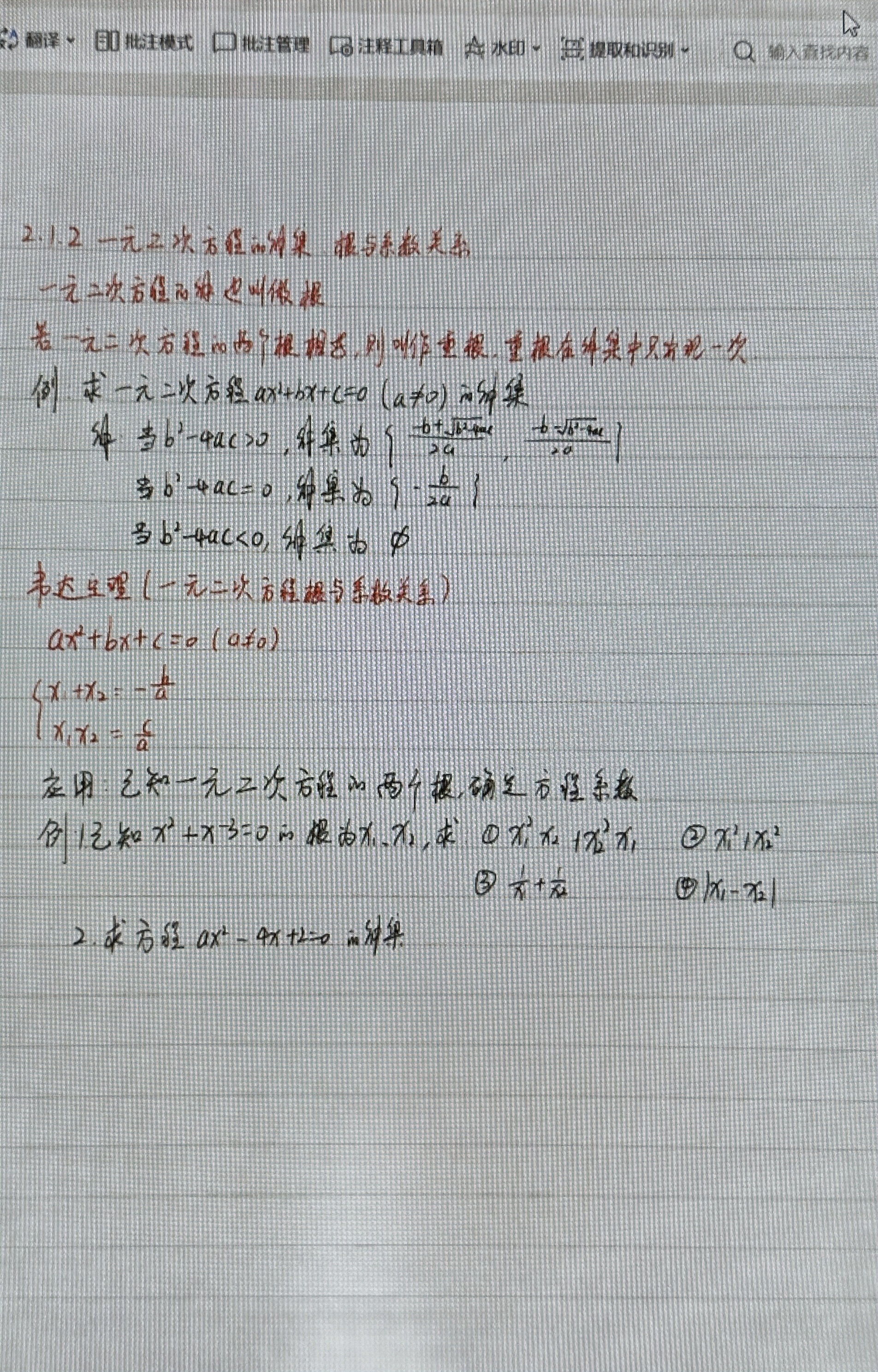}
    \caption{$s_\text{overall}=2.928$}
    \label{fig:image13}
    \end{subfigure}
    \hfill
    \begin{subfigure}[b]{0.195\textwidth}
    \centering
    \includegraphics[width=\linewidth, height=3cm, keepaspectratio]{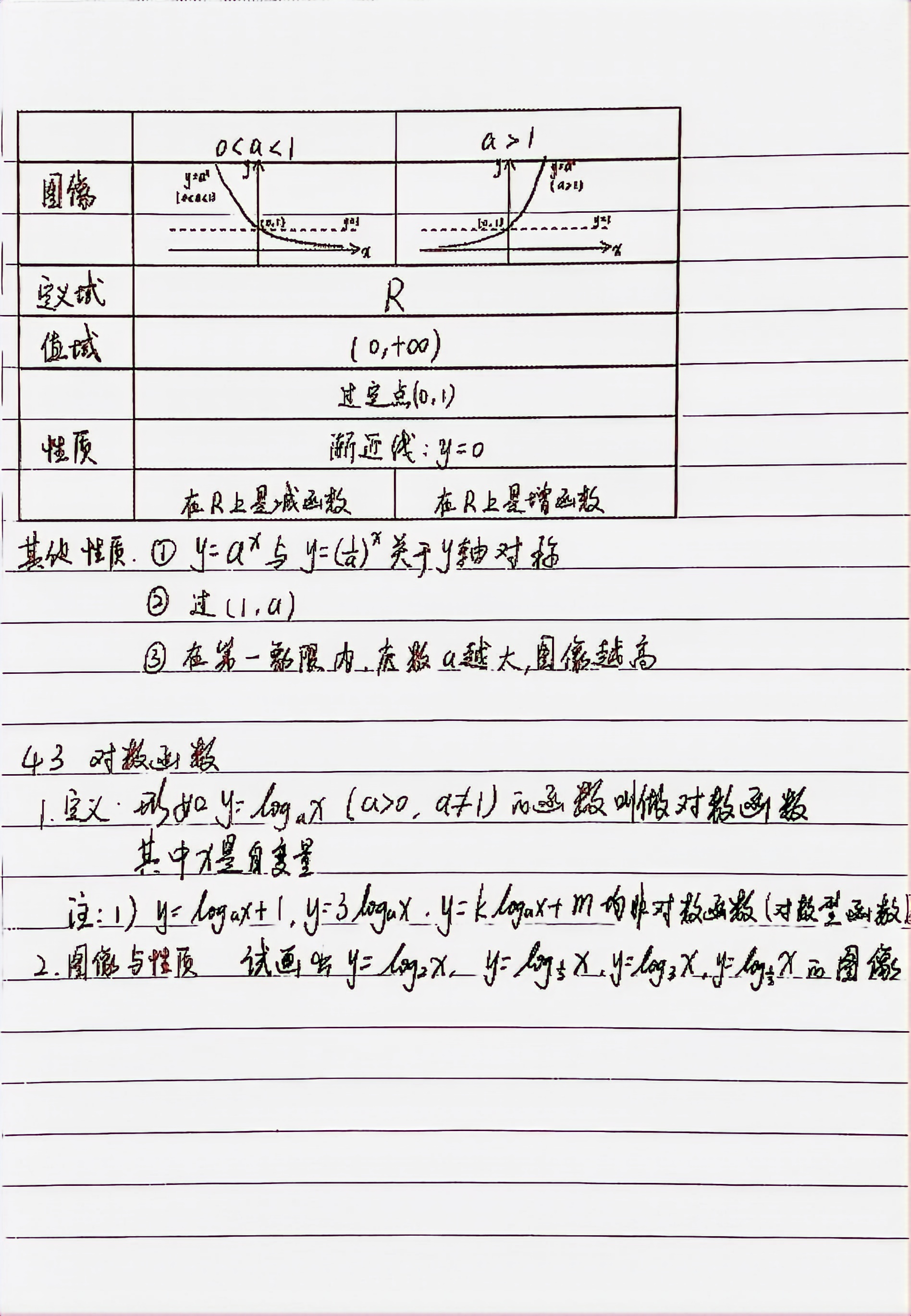}
    \caption{$s_\text{overall}=3.873$}
    \label{fig:image14}
    \end{subfigure}
    \hfill
    \begin{subfigure}[b]{0.195\textwidth}
    \centering
    \includegraphics[width=\linewidth, height=3cm, keepaspectratio]{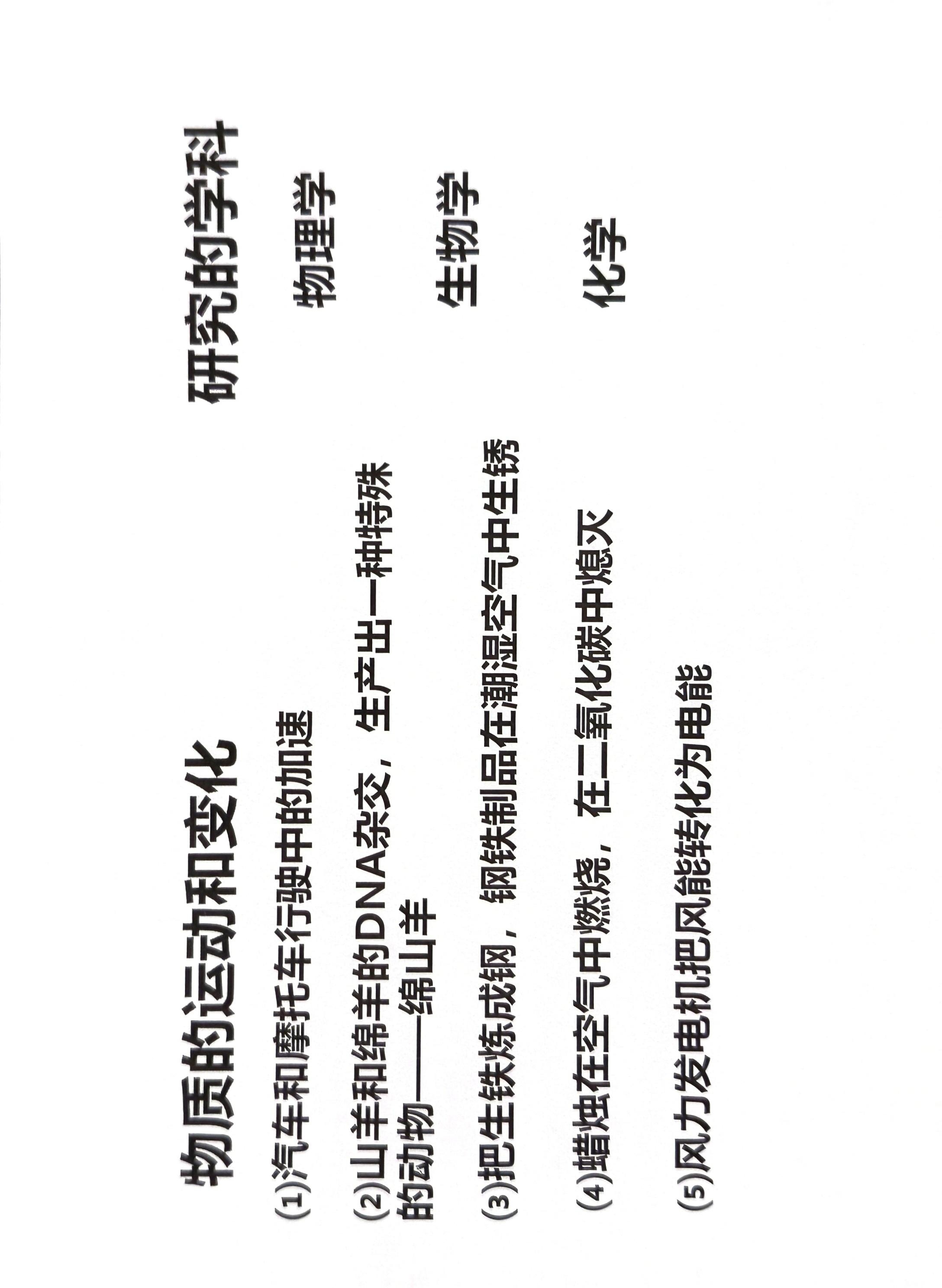}
    \caption{$s_\text{overall}=4.787$}
    \label{fig:image15}
    \end{subfigure}
    \caption{\textbf{Qualitative results} of our \method on the DIQA-5000 dataset. From left to right, each column is roughly the ``bad'', ``poor'', ``fair'', ``good'', ``excellent'' documents in sequence.}
    \vspace{-10pt}
    \label{fig:demo}
\end{figure*}

\begin{table*}[t]
\setlength\tabcolsep{8pt}
\centering
\small
\caption{
    \textbf{Ablation study of the base method} (Q-Align \vs DeQA-Score) on the DIQA-5000 val dataset. 
}
\vspace{-5pt}
\label{tab:ablation_qalign}
\begin{tabular}{c|ccc|c}
\toprule
Base Model & Sharpness & Color Fidelity & Overall & Final Score \\
\midrule
Q-Align~\cite{qalign} & 0.8604 & 0.8330 & 0.8578 & 0.8523 \\
DeQA-Score~\cite{deqa_score} (w/o fidelity loss) & 0.8711 & 0.8357 & 0.8726 & 0.8630 \\
DeQA-Score~\cite{deqa_score} (with fidelity loss) & \textbf{0.8909} & \textbf{0.8724} & \textbf{0.8882} & \textbf{0.8849} \\
\bottomrule
\end{tabular}
\vspace{10pt}
\setlength\tabcolsep{15pt}
\centering
\small
\caption{
    \textbf{Ablation study of image resolution} on the DIQA-5000 val dataset. 
}
\vspace{-5pt}
\label{tab:ablation_size}
\begin{tabular}{c|ccc|c}
\toprule
Image resolution & Sharpness & Color Fidelity & Overall & Final Score \\
\midrule
448 $\times$ 448 & 0.8908 & 0.8723 & 0.8882 & 0.8849 \\
1024 $\times$ 1024 & \textbf{0.9063} & \textbf{0.8767} & \textbf{0.9063} & \textbf{0.8989} \\
1536 $\times$ 1536 & 0.9062 & 0.8702 & 0.8839 & 0.8861 \\
\bottomrule
\end{tabular}
\vspace{10pt}
\setlength\tabcolsep{13.5pt}
\centering
\small
\caption{
    \textbf{Ablation study of different soft label construction} on the DIQA-5000 val dataset. 
}
\vspace{-5pt}
\label{tab:ablation_label}
\begin{tabular}{c|ccc|c}
\toprule
Soft Label & Sharpness & Color Fidelity & Overall & Final Score \\
\midrule
Linear Interpolation & 0.8798 & 0.8767 & 0.8644 & 0.8714 \\
Pseudo Variance & \textbf{0.9063} & \textbf{0.8767} & \textbf{0.9063} & \textbf{0.8989} \\
\bottomrule
\end{tabular}
\vspace{10pt}
\setlength\tabcolsep{14.5pt}
\centering
\small
\caption{
    \textbf{Ablation study of full tuning and LoRA tuning} on the DIQA-5000 val dataset. 
}
\vspace{-5pt}
\label{tab:ablation_tune}
\begin{tabular}{c|ccc|c}
\toprule
Finetuning method & Sharpness & Color Fidelity & Overall & Final Score \\
\midrule
Full & 0.9063 & 0.8767 & \textbf{0.9063} & 0.8989 \\
LoRA & \textbf{0.9151} & \textbf{0.8874} & 0.9054 & \textbf{0.9033} \\
\bottomrule
\end{tabular}
\vspace{-5pt}
\end{table*}

\subsection{Results in the Development Phase}\label{subsec:res}

During the development phase, we systematically explore and compare various design choices to optimize our framework.
These include alternative soft label strategies, different input resolutions, model backbones, and ensemble methods.
All experiments are tested on the validation split of the DIQA-5000 dataset, which serve as a held-out set for ablation studies and parameter tuning.
The goal is to identify the most effective configurations before finalizing the submission for the test phase.
Below, we report key findings from this tuning process and analyze the impact of each design choice.

\vspace{2pt}\noindent\textbf{Ablation study on the base method}. 
To identify a stronger baseline for our score regression task, we compare two internal methods: Q-Align~\cite{qalign} and DeQA-Score~\cite{deqa_score}, as shown in \cref{tab:ablation_qalign}. 
Both methods take mPLUG-Owl2-7B~\cite{mplugowl2} as the base model and resize the input images to $448 \times 448$, and both models are fully fine-tuned. 
The soft label in DeQA-Score is calculated by pseudo variance. 
Both methods are trained under the same settings, but DeQA-Score introduces additional design choices to improve alignment with perceptual attributes. 
Experimental results demonstrate that DeQA-Score consistently outperforms Q-Align across sharpness, color fidelity, and overall score. 
Moreover, adding the fidelity loss~\cite{deqa_score} further boosts the performance of DeQA-Score (Final Score: 0.8849 \vs 0.8523 for Q-Align), highlighting the benefit of explicitly modeling attribute-level supervision during training.

\vspace{2pt}\noindent\textbf{Ablation study on image resolution}.
To accommodate high-resolution document images, we remove the absolute position embeddings from the CLIP-based vision encoder in mPLUG-Owl2-7B. 
This modification relaxes the strict resolution constraint (typically $448 \times 448$) and allows our model to flexibly process document images of arbitrary sizes.
Leveraging this capability, we conduct an ablation study to investigate the impact of input resolution on quality prediction performance.
All models are based on mPLUG-Owl2-7B, fully fine-tuned with soft label calculated by pseudo variance. 
As shown in \cref{tab:ablation_size}, increasing the resolution from 448 to 1024 leads to notable improvements across all evaluation metrics, with the final score rising from 0.8630 to 0.8989. 
However, further increasing the resolution to 1536 does not yield additional gains and slightly decreases performance (Final Score: 0.8861), possibly due to overfitting or increased noise from redundant details.
These results confirm that the resolution flexibility enabled by removing absolute position embeddings is essential for handling diverse document layouts, and a resolution of 1024 provides the optimal balance between information preservation and computational cost.

\vspace{2pt}\noindent\textbf{Ablation study on soft label designs}.
We compare two soft label construction strategies, linear interpolation and pseudo variance, to assess their impact on quality regression performance. 
Both models in \cref{tab:ablation_label} use the same DeQA-Score variant and mPLUG-Owl2-7B as the base model.
The image resolution is fixed to $1024 \times 1024$, and the model is fully fine-tuned. 
As shown in \cref{tab:ablation_label}, both methods share the same color fidelity score (0.8767), but pseudo variance yields significantly better performance in sharpness (0.9063 vs. 0.8798), overall quality (0.9063 vs. 0.8644), and final score (0.8989 vs. 0.8714). 
These results highlight the advantage of the pseudo variance design, which better captures the perceptual uncertainty and diversity inherent in human annotations, leading to more reliable and discriminative supervision signals.

\vspace{2pt}\noindent\textbf{Ablation study on finetuning strategies}.
We further investigate the effect of different finetuning strategies using the DeQA-Score variant and mPLUG-Owl2-7B as the base model. The image resolution is fixed to $1024 \times 1024$. 
Default soft label is obtained through pseudo variance. 
As shown in~\cref{tab:ablation_tune}, both full finetuning and LoRA achieve competitive results on the DIQA-5000 validation set. 
Notably, LoRA yields slightly better performance across all attributes, with improvements in sharpness (0.9151 \vs 0.9063), color fidelity (0.8874 \vs 0.8767), and final score (0.9033 \vs 0.8989). 
These results demonstrate that LoRA can serve as an efficient and effective alternative to full finetuning, achieving comparable or even superior quality regression performance with significantly fewer trainable parameters.

\subsection{Results in the Test Phase}

\begin{table*}[t]
\setlength\tabcolsep{6pt}
\centering
\small
\caption{
    \textbf{Score regression results with mPLUG-Owl2 as base model} on the DIQA-5000 test dataset. 
}
\vspace{-5pt}
\label{tab:mplug}
\begin{tabular}{ccccc|ccc|c}
\toprule
\# & Base Model & Pretrain & Resolution & Finetune & Sharpness & Color Fidelity & Overall & Final Score \\
\midrule
m0 & mPLUG-Owl2-7B & - & 1024 & Full & \textbf{0.9106} & 0.8924 & \textbf{0.9182} & 0.9098 \\
m1 & mPLUG-Owl2-7B & - & 1024 & LoRA & 0.9099 & \textbf{0.9062} & 0.9135 & 0.9108 \\
m2 & mPLUG-Owl2-7B & - & 1536 & Full & 0.8307 & 0.8960 & 0.8897 & 0.8765 \\
m3 & mPLUG-Owl2-7B & KonIQ & 1024 & LoRA & 0.9056 & 0.9048 & 0.9172 & \textbf{0.9112} \\
\bottomrule
\end{tabular}
\vspace{10pt}
\setlength\tabcolsep{4pt}
\centering
\small
\caption{
    \textbf{Score regression results with Qwen2.5-VL as base model} on the DIQA-5000 test dataset. 
}
\vspace{-5pt}
\label{tab:qwen}
\begin{tabular}{ccccc|ccc|c}
\toprule
\# & Base Model & Ensemble Method & Resolution & Finetune & Sharpness & Color Fidelity & Overall & Final Score \\
\midrule
Q0 & Qwen2.5-VL-7B & - & Original & Full & 0.9067 & 0.8947 & 0.9100 & 0.9054 \\
Q1 & Qwen2.5-VL-7B & 5-fold & Original & Full & \textbf{0.9175} & \textbf{0.9185} & \textbf{0.9289} & \textbf{0.9235} \\
\bottomrule
\end{tabular}
\vspace{10pt}
\setlength\tabcolsep{10.5pt}
\centering
\small
\caption{
    \textbf{Score regression results after ensemble method} on the DIQA-5000 test dataset. 
}
\vspace{-5pt}
\label{tab:ensemble}
\begin{tabular}{cc|ccc|c}
\toprule
Model Indices & Ensemble Method & Sharpness & Color Fidelity & Overall & Final Score \\
\midrule
m0 & Prompt & 0.9106 & 0.8926 & 0.9182 & 0.9099 \\
m0 + m1 + m3 & Average & 0.9145 & 0.9089 & 0.9232 & 0.9174 \\
m0 + m1 + m3 + Q0 & Average & 0.9213 & 0.9149 & 0.9292 & 0.9234 \\
m0 + m1 + m3 + Q0 + Q1 & Average & \textbf{0.9275} & \textbf{0.9198} & \textbf{0.9339} & \textbf{0.9288} \\
\bottomrule
\end{tabular}
\vspace{-5pt}
\end{table*}

In the test phase, we evaluate our system on the hidden test set of the DIQA-5000 benchmark.
We select the best-performing configurations identified during the development phase, including the optimal input resolution, soft label strategy, and model backbone.
To further enhance performance, we apply both model ensemble and prompt ensemble techniques, which prove effective in reducing prediction variance and improving robustness.
By averaging outputs across different models and multiple prompt templates, our final system achieves strong generalization on the unseen test set, validating the effectiveness of the proposed design choices.

\vspace{2pt}\noindent\textbf{Score regression results with mPLUG-Owl2 as the base model}. 
\cref{tab:mplug} presents the score regression results on the DIQA-5000 test set using mPLUG-Owl2-7B as the base model under different configurations.
We evaluate the effects of fine-tuning strategy, input resolution, and pre-training on natural image quality data (KonIQ~\cite{koniq}). 
Among the models without pretraining (from m0 to m2), using a resolution of 1024 achieves better performance than 1536, confirming our earlier observation that 1024 offers the best trade-off between visual detail and stability. 
It is consistent with our observations on the development phase that LoRA tuning (m1) achieves comparable or slightly better performance than full finetuning (m0), with a higher final score (0.9108 vs. 0.9098), while being more parameter-efficient. 
Moreover, pretraining on KonIQ (m3) further improves the performance, achieving the best final score of 0.9112, suggesting that prior knowledge from natural image quality assessment can benefit document quality assessment tasks. 
Overall, these results validate the robustness of our framework across different training setups and highlight the effectiveness of lightweight adaptation and cross-domain pretraining.

\vspace{2pt}\noindent\textbf{Score regression results with Qwen2.5-VL as the base model}. 
\cref{tab:qwen} reports the score regression results on the DIQA-5000 test set using Qwen2.5-VL-7B as the base model. 
A key advantage of Qwen2.5-VL lies in its ability to accept document images at their original resolution, avoiding aggressive resizing that might otherwise degrade fine-grained layout or text features. 
In Q0, a single model trained with full finetuning achieves a Final Score of 0.9054, showing solid performance across sharpness, color fidelity, and overall quality. 
To further improve robustness, we apply a 5-fold ensemble strategy in Q1, averaging predictions from five independently trained models. 
This ensemble leads to a substantial performance gain, resulting in a substantial performance gain, with the Final Score increasing to 0.9235 and noticeable improvements in both sharpness and color fidelity. 
These results highlight the strength of Qwen2.5-VL in document quality assessment, particularly when combined with ensemble techniques that reduce prediction variance and boost generalization.

\vspace{2pt}\noindent\textbf{Score regression results with different ensemble strategies}. 
In \cref{tab:ensemble}, we summarize the impact of ensemble strategies on score regression performance. 
We begin with prompt ensemble on a single model (m0), which yields a Final Score of 0.9099, the same as original 0.9098 (m0), showing almost no performance improvement. 
Then, we combine multiple variants of mPLUG-Owl2-7B (m0 + m1 + m3) via averaging, resulting in a notable performance gain across all metrics, with the Final Score improving to 0.9174. 
Next, we incorporate Qwen2.5-VL (Q0) into the ensemble, further boosting the Final Score to 0.9234, demonstrating the complementarity of different MLLM architectures. 
Finally, the full ensemble, including both mPLUG-Owl2 and Qwen2.5-VL models (m0 + m1 + m3 + Q0 + Q1), achieves the best overall results, with a Final Score of 0.9288 and strong performance improvements in sharpness (0.9275) and color fidelity (0.9198). 
These results validate the effectiveness of model-level ensemble strategies in improving the robustness and accuracy of document quality assessment.

\vspace{2pt}\noindent\textbf{Qualitative results.} 
\cref{fig:demo} presents representative prediction results on the DIQA-5000 dataset, illustrating the model’s ability to align well with human-perceived document quality. 
From left to right, each column roughly corresponds to increasing quality levels: ``bad'', ``poor'', ``fair'', ``good'', and ``excellent''. 
Low-scoring examples (\eg, (a), (f), and (k)) exhibit severe degradation such as extreme blur, noise, or compression artifacts, which significantly impact readability. 
Mid-range samples (\eg, (c), (h), and (m)) are generally legible but suffer from issues like moiré patterns, uneven lighting, or low contrast. 
High-quality predictions (\eg, (e), (j), and (o)) show clean layouts, sharp text, and consistent formatting. 
The predicted scores increase consistently with the perceptual quality, demonstrating that our \method model effectively captures fine-grained differences in document quality.

\section{Conclusions}\label{sec:conclusion}

In this work, we present \method, an effective MLLM-based framework for document quality assessment that builds upon the success of DeQA-Score in image quality scoring. 
By adapting the model to handle document-specific characteristics and proposing two alternative soft label strategies to address missing variance annotations in the DIQA-5000 dataset, we enable continuous quality score regression within a unified visual-language architecture. 
We also relax the resolution constrains to support the large resolution of document images, and introduce prompt ensemble and model ensemble methods to further enhance the performance. 
Our method demonstrates strong performance across diverse document types and degradation patterns, outperforming existing baselines in accuracy and generalizability. 
We believe this work offers a promising direction for future research in quality-aware document processing.

{
    \small
    \bibliographystyle{ieeenat_fullname}
    \bibliography{main}

\begin{thebibliography}{52}
\providecommand{\natexlab}[1]{#1}
\providecommand{\url}[1]{\texttt{#1}}
\expandafter\ifx\csname urlstyle\endcsname\relax
  \providecommand{\doi}[1]{doi: #1}\else
  \providecommand{\doi}{doi: \begingroup \urlstyle{rm}\Url}\fi

\bibitem[Alaei(2019)]{alaei2019new}
Alireza Alaei.
\newblock A new document image quality assessment method based on hast derivations.
\newblock In \emph{International Conference on Document Analysis and Recognition (ICDAR)}, 2019.

\bibitem[Alaei et~al.(2015)Alaei, Conte, and Raveaux]{alaei2015document}
Alireza Alaei, Donatello Conte, and Romain Raveaux.
\newblock Document image quality assessment based on improved gradient magnitude similarity deviation.
\newblock In \emph{International Conference on Document Analysis and Recognition (ICDAR)}, 2015.

\bibitem[Alaei et~al.(2023)Alaei, Bui, Doermann, and Pal]{alaei2023document}
Alireza Alaei, Vinh Bui, David Doermann, and Umapada Pal.
\newblock Document image quality assessment: A survey.
\newblock \emph{ACM Computing Surveys}, 2023.

\bibitem[Bai et~al.(2025)Bai, Chen, Liu, Wang, Ge, Song, Dang, Wang, Wang, Tang, Zhong, Zhu, Yang, Li, Wan, Wang, Ding, Fu, Xu, Ye, Zhang, Xie, Cheng, Zhang, Yang, Xu, and Lin]{qwen2.5vl}
Shuai Bai, Keqin Chen, Xuejing Liu, Jialin Wang, Wenbin Ge, Sibo Song, Kai Dang, Peng Wang, Shijie Wang, Jun Tang, Humen Zhong, Yuanzhi Zhu, Mingkun Yang, Zhaohai Li, Jianqiang Wan, Pengfei Wang, Wei Ding, Zheren Fu, Yiheng Xu, Jiabo Ye, Xi Zhang, Tianbao Xie, Zesen Cheng, Hang Zhang, Zhibo Yang, Haiyang Xu, and Junyang Lin.
\newblock {Qwen2.5-VL} technical report.
\newblock \emph{arXiv preprint arXiv:2502.13923}, 2025.

\bibitem[Chen et~al.(2024)Chen, Sensen, Wu, Liao, Zhang, Wang, Sun, Yan, and Lin]{qground}
Chaofeng Chen, Yang Sensen, Haoning Wu, Liang Liao, Zicheng Zhang, Annan Wang, Wenxiu Sun, Qiong Yan, and Weisi Lin.
\newblock {Q-Ground}: Image quality grounding with large multi-modality models.
\newblock In \emph{ACM MM}, 2024.

\bibitem[Chen et~al.(2025)Chen, Wu, Ma, and Zhang]{afine}
Du Chen, Tianhe Wu, Kede Ma, and Lei Zhang.
\newblock Toward generalized image quality assessment: Relaxing the perfect reference quality assumption.
\newblock In \emph{CVPR}, 2025.

\bibitem[Ding et~al.(2020)Ding, Ma, Wang, and Simoncelli]{dists}
Keyan Ding, Kede Ma, Shiqi Wang, and Eero~P Simoncelli.
\newblock Image quality assessment: Unifying structure and texture similarity.
\newblock \emph{IEEE TPAMI}, 2020.

\bibitem[Ding et~al.(2021)Ding, Liu, Zou, Wang, and Ma]{A-DISTS}
Keyan Ding, Yi Liu, Xueyi Zou, Shiqi Wang, and Kede Ma.
\newblock Locally adaptive structure and texture similarity for image quality assessment.
\newblock In \emph{ACM MM}, 2021.

\bibitem[Fang et~al.(2020)Fang, Zhu, Zeng, Ma, and Wang]{spaq}
Yuming Fang, Hanwei Zhu, Yan Zeng, Kede Ma, and Zhou Wang.
\newblock Perceptual quality assessment of smartphone photography.
\newblock In \emph{CVPR}, 2020.

\bibitem[Ghildyal and Liu(2022)]{ghildyal2022stlpips}
Abhijay Ghildyal and Feng Liu.
\newblock Shift-tolerant perceptual similarity metric.
\newblock In \emph{ECCV}, 2022.

\bibitem[Hosu et~al.(2020)Hosu, Lin, Sziranyi, and Saupe]{koniq}
Vlad Hosu, Hanhe Lin, Tamas Sziranyi, and Dietmar Saupe.
\newblock {KonIQ-10K}: An ecologically valid database for deep learning of blind image quality assessment.
\newblock \emph{IEEE TIP}, 2020.

\bibitem[Hu et~al.(2021)Hu, Shen, Wallis, Allen-Zhu, Li, Wang, Wang, and Chen]{lora}
Edward~J Hu, Yelong Shen, Phillip Wallis, Zeyuan Allen-Zhu, Yuanzhi Li, Shean Wang, Lu Wang, and Weizhu Chen.
\newblock {LoRA}: Low-rank adaptation of large language models.
\newblock In \emph{ICLR}, 2021.

\bibitem[Ilya and Frank(2019)]{adamw}
Loshchilov Ilya and Hutter Frank.
\newblock Decoupled weight decay regularization.
\newblock In \emph{ICLR}, 2019.

\bibitem[Jinjin et~al.(2020)Jinjin, Haoming, Haoyu, Xiaoxing, Ren, and Chao]{pipal}
Gu Jinjin, Cai Haoming, Chen Haoyu, Ye Xiaoxing, Jimmy~S Ren, and Dong Chao.
\newblock {PIPAL}: a large-scale image quality assessment dataset for perceptual image restoration.
\newblock In \emph{ECCV}, 2020.

\bibitem[Kang et~al.(2014)Kang, Ye, Li, and Doermann]{diqa_deep}
Le Kang, Peng Ye, Yi Li, and David Doermann.
\newblock A deep learning approach to document image quality assessment.
\newblock In \emph{International Conference on Image Processing (ICIP)}, 2014.

\bibitem[Ke et~al.(2021)Ke, Wang, Wang, Milanfar, and Yang]{musiq}
Junjie Ke, Qifei Wang, Yilin Wang, Peyman Milanfar, and Feng Yang.
\newblock {MUSIQ}: Multi-scale image quality transformer.
\newblock In \emph{CVPR}, 2021.

\bibitem[Kumar et~al.(2012)Kumar, Chen, and Doermann]{Kumar2012Sharpness}
J. Kumar, F. Chen, and David Doermann.
\newblock Sharpness estimation for document and scene images.
\newblock In \emph{ICPR}, 2012.

\bibitem[Li et~al.(2018)Li, Zhu, and Qiu]{cg_diqa}
Hongyu Li, Fan Zhu, and Junhua Qiu.
\newblock {CG-DIQA}: No-reference document image quality assessment based on character gradient.
\newblock In \emph{International Conference on Pattern Recognition (ICPR)}, 2018.

\bibitem[Lin et~al.(2019)Lin, Hosu, and Saupe]{kadid}
Hanhe Lin, Vlad Hosu, and Dietmar Saupe.
\newblock {KADID-10K}: A large-scale artificially distorted iqa database.
\newblock In \emph{International Conference on Quality of Multimedia Experience (QoMEX)}, 2019.

\bibitem[Liu et~al.(2024)Liu, Li, Wu, and Lee]{llava}
Haotian Liu, Chunyuan Li, Qingyang Wu, and Yong~Jae Lee.
\newblock Visual instruction tuning.
\newblock In \emph{NeurIPS}, 2024.

\bibitem[Lu and Dooms(2019)]{lu2019deep}
Tan Lu and Ann Dooms.
\newblock A deep transfer learning approach to document image quality assessment.
\newblock In \emph{International Conference on Document Analysis and Recognition (ICDAR)}, 2019.

\bibitem[Ma et~al.(2017)Ma, Yang, Yang, and Yang]{ma2017learning}
Chao Ma, Chih-Yuan Yang, Xiaokang Yang, and Ming-Hsuan Yang.
\newblock Learning a no-reference quality metric for single-image super-resolution.
\newblock \emph{Computer Vision and Image Understanding}, 2017.

\bibitem[Min et~al.(2017)Min, Ma, Gu, Zhai, Wang, and Lin]{min2017unified}
Xiongkuo Min, Kede Ma, Ke Gu, Guangtao Zhai, Zhou Wang, and Weisi Lin.
\newblock Unified blind quality assessment of compressed natural, graphic, and screen content images.
\newblock \emph{IEEE TIP}, 2017.

\bibitem[Mittal et~al.(2012)Mittal, Soundararajan, and Bovik]{niqe}
Anish Mittal, Rajiv Soundararajan, and Alan~C Bovik.
\newblock Making a “completely blind” image quality analyzer.
\newblock \emph{IEEE Sign. Process. Letters}, 2012.

\bibitem[Ouyang et~al.(2022)Ouyang, Wu, Jiang, Almeida, Wainwright, Mishkin, Zhang, Agarwal, Slama, Ray, et~al.]{gpt3.5}
Long Ouyang, Jeffrey Wu, Xu Jiang, Diogo Almeida, Carroll Wainwright, Pamela Mishkin, Chong Zhang, Sandhini Agarwal, Katarina Slama, Alex Ray, et~al.
\newblock Training language models to follow instructions with human feedback.
\newblock In \emph{NeurIPS}, 2022.

\bibitem[Prashnani et~al.(2018)Prashnani, Cai, Mostofi, and Sen]{pieapp}
Ekta Prashnani, Hong Cai, Yasamin Mostofi, and Pradeep Sen.
\newblock {PieAPP}: Perceptual image-error assessment through pairwise preference.
\newblock In \emph{CVPR}, 2018.

\bibitem[Radford et~al.(2021)Radford, Kim, Hallacy, Ramesh, Goh, Agarwal, Sastry, Askell, Mishkin, Clark, et~al.]{clip}
Alec Radford, Jong~Wook Kim, Chris Hallacy, Aditya Ramesh, Gabriel Goh, Sandhini Agarwal, Girish Sastry, Amanda Askell, Pamela Mishkin, Jack Clark, et~al.
\newblock Learning transferable visual models from natural language supervision.
\newblock In \emph{ICML}, 2021.

\bibitem[Saad et~al.(2012)Saad, Bovik, and Charrier]{saad2012blind}
Michele~A Saad, Alan~C Bovik, and Christophe Charrier.
\newblock Blind image quality assessment: A natural scene statistics approach in the dct domain.
\newblock \emph{IEEE TIP}, 2012.

\bibitem[Schulz et~al.(2022)Schulz, Maureira, Tapia, and Busch]{schulz2022identity}
Daniel Schulz, Jose Maureira, Juan Tapia, and Christoph Busch.
\newblock Identity documents image quality assessment.
\newblock In \emph{European Signal Processing Conference (EUSIPCO)}, 2022.

\bibitem[Series(2012)]{five_level}
B Series.
\newblock Methodology for the subjective assessment of the quality of television pictures.
\newblock \emph{Recommendation ITU-R BT}, 2012.

\bibitem[Shemiakina et~al.(2021)Shemiakina, Limonova, Skoryukina, Arlazarov, and Nikolaev]{shemiakina2021method}
Julia Shemiakina, Elena Limonova, Natalya Skoryukina, Vladimir~V Arlazarov, and Dmitry~P Nikolaev.
\newblock A method of image quality assessment for text recognition on camera-captured and projectively distorted documents.
\newblock \emph{Mathematics}, 2021.

\bibitem[Touvron et~al.(2023)Touvron, Lavril, Izacard, Martinet, Lachaux, Lacroix, Rozi{\`e}re, Goyal, Hambro, Azhar, et~al.]{llama}
Hugo Touvron, Thibaut Lavril, Gautier Izacard, Xavier Martinet, Marie-Anne Lachaux, Timoth{\'e}e Lacroix, Baptiste Rozi{\`e}re, Naman Goyal, Eric Hambro, Faisal Azhar, et~al.
\newblock {LLaMA}: Open and efficient foundation language models.
\newblock \emph{arXiv preprint arXiv:2302.13971}, 2023.

\bibitem[Wang et~al.(2023)Wang, Chan, and Loy]{clipiqa}
Jianyi Wang, Kelvin~CK Chan, and Chen~Change Loy.
\newblock Exploring clip for assessing the look and feel of images.
\newblock In \emph{AAAI}, 2023.

\bibitem[Wang et~al.(2004)Wang, Bovik, Sheikh, and Simoncelli]{ssim}
Zhou Wang, Alan~C Bovik, Hamid~R Sheikh, and Eero~P Simoncelli.
\newblock Image quality assessment: from error visibility to structural similarity.
\newblock \emph{IEEE TIP}, 2004.

\bibitem[Wu et~al.(2024{\natexlab{a}})Wu, Zhang, Zhang, Chen, Liao, Wang, Li, Sun, Yan, Zhai, et~al.]{qbench}
Haoning Wu, Zicheng Zhang, Erli Zhang, Chaofeng Chen, Liang Liao, Annan Wang, Chunyi Li, Wenxiu Sun, Qiong Yan, Guangtao Zhai, et~al.
\newblock {Q-Bench}: A benchmark for general-purpose foundation models on low-level vision.
\newblock In \emph{ICLR}, 2024{\natexlab{a}}.

\bibitem[Wu et~al.(2024{\natexlab{b}})Wu, Zhang, Zhang, Chen, Liao, Wang, Xu, Li, Hou, Zhai, et~al.]{qinstruct}
Haoning Wu, Zicheng Zhang, Erli Zhang, Chaofeng Chen, Liang Liao, Annan Wang, Kaixin Xu, Chunyi Li, Jingwen Hou, Guangtao Zhai, et~al.
\newblock {Q-Instruct}: Improving low-level visual abilities for multi-modality foundation models.
\newblock In \emph{CVPR}, 2024{\natexlab{b}}.

\bibitem[Wu et~al.(2024{\natexlab{c}})Wu, Zhang, Zhang, Chen, Liao, Li, Gao, Wang, Zhang, Sun, et~al.]{qalign}
Haoning Wu, Zicheng Zhang, Weixia Zhang, Chaofeng Chen, Liang Liao, Chunyi Li, Yixuan Gao, Annan Wang, Erli Zhang, Wenxiu Sun, et~al.
\newblock {Q-Align}: Teaching {LMMs} for visual scoring via discrete text-defined levels.
\newblock In \emph{ICML}, 2024{\natexlab{c}}.

\bibitem[Wu et~al.(2024{\natexlab{d}})Wu, Zhu, Zhang, Zhang, Chen, Liao, Li, Wang, Sun, Yan, et~al.]{coinstruct}
Haoning Wu, Hanwei Zhu, Zicheng Zhang, Erli Zhang, Chaofeng Chen, Liang Liao, Chunyi Li, Annan Wang, Wenxiu Sun, Qiong Yan, et~al.
\newblock Towards open-ended visual quality comparison.
\newblock In \emph{ECCV}, 2024{\natexlab{d}}.

\bibitem[Wu et~al.(2024{\natexlab{e}})Wu, Ma, Liang, Yang, and Zhang]{iqasurvey_tianhe}
Tianhe Wu, Kede Ma, Jie Liang, Yujiu Yang, and Lei Zhang.
\newblock A comprehensive study of multimodal large language models for image quality assessment.
\newblock In \emph{ECCV}, 2024{\natexlab{e}}.

\bibitem[Xu et~al.(2016)Xu, Ye, Li, Liu, and Doermann]{xu2016no}
Jingtao Xu, Peng Ye, Qiaohong Li, Yong Liu, and David Doermann.
\newblock No-reference document image quality assessment based on high order image statistics.
\newblock In \emph{IEEE International Conference on Image Processing (ICIP)}, 2016.

\bibitem[Yang et~al.(2022)Yang, Wu, Shi, Lao, Gong, Cao, Wang, and Yang]{maniqa}
Sidi Yang, Tianhe Wu, Shuwei Shi, Shanshan Lao, Yuan Gong, Mingdeng Cao, Jiahao Wang, and Yujiu Yang.
\newblock {MANIQA}: Multi-dimension attention network for no-reference image quality assessment.
\newblock In \emph{CVPRW}, 2022.

\bibitem[Ye and Doermann(2013)]{diqa_survey}
Peng Ye and David Doermann.
\newblock Document image quality assessment: A brief survey.
\newblock In \emph{International Conference on Document Analysis and Recognition}, 2013.

\bibitem[Ye et~al.(2024)Ye, Xu, Ye, Yan, Hu, Liu, Qian, Zhang, and Huang]{mplugowl2}
Qinghao Ye, Haiyang Xu, Jiabo Ye, Ming Yan, Anwen Hu, Haowei Liu, Qi Qian, Ji Zhang, and Fei Huang.
\newblock {mPLUG-Owl2}: Revolutionizing multi-modal large language model with modality collaboration.
\newblock In \emph{CVPR}, 2024.

\bibitem[You et~al.(2024{\natexlab{a}})You, Gu, Li, Cai, Zhu, Dong, and Xue]{depictqav2}
Zhiyuan You, Jinjin Gu, Zheyuan Li, Xin Cai, Kaiwen Zhu, Chao Dong, and Tianfan Xue.
\newblock Descriptive image quality assessment in the wild.
\newblock \emph{arXiv preprint arXiv:2405.18842}, 2024{\natexlab{a}}.

\bibitem[You et~al.(2024{\natexlab{b}})You, Li, Gu, Yin, Xue, and Dong]{depictqa}
Zhiyuan You, Zheyuan Li, Jinjin Gu, Zhenfei Yin, Tianfan Xue, and Chao Dong.
\newblock Depicting beyond scores: Advancing image quality assessment through multi-modal language models.
\newblock In \emph{ECCV}, 2024{\natexlab{b}}.

\bibitem[You et~al.(2025)You, Cai, Gu, Xue, and Dong]{deqa_score}
Zhiyuan You, Xin Cai, Jinjin Gu, Tianfan Xue, and Chao Dong.
\newblock Teaching large language models to regress accurate image quality scores using score distribution.
\newblock In \emph{CVPR}, 2025.

\bibitem[Zhang et~al.(2011)Zhang, Zhang, Mou, and Zhang]{fsim}
Lin Zhang, Lei Zhang, Xuanqin Mou, and David Zhang.
\newblock {FSIM}: A feature similarity index for image quality assessment.
\newblock \emph{IEEE TIP}, 2011.

\bibitem[Zhang et~al.(2024)Zhang, Wang, and Wan]{zhang2024efficient}
Lina Zhang, Kaiyuan Wang, and Yi Wan.
\newblock An efficient transformer--cnn network for document image binarization.
\newblock \emph{Electronics}, 2024.

\bibitem[Zhang et~al.(2018)Zhang, Isola, Efros, Shechtman, and Wang]{lpips}
Richard Zhang, Phillip Isola, Alexei~A Efros, Eli Shechtman, and Oliver Wang.
\newblock The unreasonable effectiveness of deep features as a perceptual metric.
\newblock In \emph{CVPR}, 2018.

\bibitem[Zhang et~al.(2021)Zhang, Ma, Zhai, and Yang]{unique}
Weixia Zhang, Kede Ma, Guangtao Zhai, and Xiaokang Yang.
\newblock Uncertainty-aware blind image quality assessment in the laboratory and wild.
\newblock \emph{IEEE TIP}, 2021.

\bibitem[Zhang et~al.(2023)Zhang, Zhai, Wei, Yang, and Ma]{liqe}
Weixia Zhang, Guangtao Zhai, Ying Wei, Xiaokang Yang, and Kede Ma.
\newblock Blind image quality assessment via vision-language correspondence: A multitask learning perspective.
\newblock In \emph{CVPR}, 2023.

\bibitem[Zhu et~al.(2024)Zhu, Wu, Li, Zhang, Chen, Zhu, Fang, Zhai, Lin, and Wang]{compare2score}
Hanwei Zhu, Haoning Wu, Yixuan Li, Zicheng Zhang, Baoliang Chen, Lingyu Zhu, Yuming Fang, Guangtao Zhai, Weisi Lin, and Shiqi Wang.
\newblock Adaptive image quality assessment via teaching large multimodal model to compare.
\newblock \emph{arXiv preprint arXiv:2405.19298}, 2024.

\end{thebibliography}
}

\end{document}